\documentclass[10pt,journal,compsoc]{IEEEtran}

\ifCLASSOPTIONcompsoc
  \usepackage[nocompress]{cite}
\else
  \usepackage{cite}
\fi

\ifCLASSINFOpdf
  \usepackage[pdftex]{graphicx}
\else
\fi

\usepackage{amsmath}

\usepackage{array}

  \usepackage[caption=false,font=footnotesize,labelfont=sf,textfont=sf]{subfig}

\usepackage{color}

\RequirePackage[normalem]{ulem}

\newcommand{\bs}{\boldsymbol}
\newcommand{\ph}[2]{{\boldsymbol\phi}_{#1}^{#2}}

\newcommand{\SW}{{\bf{\Sigma}}_W}
\newcommand{\SB}{{\bf{\Sigma}}_B}

\begin{document}

\title{Discriminant analysis based on projection onto generalized difference subspace}

\author{Kazuhiro Fukui, ~Naoya Sogi, ~Takumi Kobayashi, ~Jing-Hao Xue, ~Atsuto Maki


\IEEEcompsocitemizethanks{\IEEEcompsocthanksitem K. Fukui and N. Sogi are with the Department of Computer Science, Graduate School of Systems and Information Engineering, University of Tsukuba, Japan.
\IEEEcompsocthanksitem T. Kobayashi is with the National Institute of Advanced Industrial Science and Technology (AIST), Japan.
\IEEEcompsocthanksitem J.-H. Xue is with the Department of Statistical Science, University College London, UK.
\IEEEcompsocthanksitem A. Maki is with the School of Electrical Engineering and Computer Science, Royal Institute of Technology (KTH), Stockholm, Sweden.}
}




\IEEEtitleabstractindextext{%
\begin{abstract}
This paper discusses a new type of discriminant analysis based on the orthogonal projection of data onto a generalized difference subspace (GDS). In our previous work, we have demonstrated that GDS projection works as the quasi-orthogonalization of class subspaces, which is an effective feature extraction for subspace based classifiers. Interestingly, GDS projection also works as a discriminant feature extraction through a similar mechanism to the Fisher discriminant analysis (FDA). A direct proof of the connection between GDS projection and FDA is difficult due to the significant difference in their formulations. To avoid the difficulty, we first introduce geometrical Fisher discriminant analysis (gFDA) based on a simplified Fisher criterion. Our simplified Fisher criterion is derived from a heuristic yet practically plausible principle: the direction of the sample mean vector of a class is in most cases almost equal to that of the first principal component vector of the class, under the condition that the principal component vectors are calculated by applying the principal component analysis (PCA) without data centering. gFDA can work stably even under few samples, bypassing the small sample size (SSS) problem of FDA. Next, we prove that gFDA is equivalent to GDS projection with a small correction term. This equivalence ensures GDS projection to inherit the discriminant ability from FDA via gFDA. Furthermore, to enhance the performances of gFDA and GDS projection, we normalize the projected vectors on the discriminant spaces. Extensive experiments using the extended Yale B+ database and the CMU face database show that gFDA and GDS projection have equivalent or better performance than the original FDA and its extensions.
\end{abstract}


}

\maketitle

\IEEEdisplaynontitleabstractindextext

\IEEEpeerreviewmaketitle

\IEEEraisesectionheading{\section{Introduction}\label{sec:introduction}}
In this paper, we discuss a new type of discriminant analysis based on a projection onto the generalized difference subspace (GDS) that represents difference among multiple class subspaces  \cite{gds}. GDS is defined as a generalization of the difference subspace (DS) that represents the difference between two subspaces. DS is a natural extension of the difference vector between two vectors. 

The orthogonal projection of data or subspaces onto a GDS, which is called GDS projection, has two natures in feature extraction. These natures can be alternatively exploited. 
%
One nature is to enlarge the angles among class subspaces to make their relationship closer to the orthogonal status \cite{gds}. As a result, GDS projection works as quasi-orthogonalization and an effective feature extraction technique for subspace based classifiers such as the subspace method and the mutual subspace method \cite{cmsm,kcmsm}. 
The other nature is to serve for discriminative feature extraction, through a mechanism similar to the Fisher discriminant analysis (FDA) \cite{fda1,fda2}. 

In this paper, we clarify the latter nature both theoretically and empirically by exploring the close connection between GDS projection and FDA. 
However, a direct proof of their close connection would not be straightforward due to the significant difference in their formulations. 
To circumvent the complication, we introduce geometrical Fisher discriminant analysis (gFDA) that is a discriminant analysis based on a simplified Fisher criterion in terms of class representation. Then, 
we indirectly prove the close connection between GDS projection and FDA via gFDA, where gFDA serves as an intermediate concept between FDA and GDS projection since gFDA inherits the intrinsic mechanism from GDS projection and the discriminant ability from FDA.
 
Our simplification starts with the introduction of a heuristic principle that the directions of the sample mean vector and the first principal component vector of a class are nearly equivalent, 
given the condition that 
the PCA without data centering (subtracting the mean) is applied to calculate the principal component vectors. 

This heuristic principle enables us to reasonably represent the original Fisher criterion using the principal component vectors and their weights (eigenvalues) of all the classes involved in the classification task.
That is, based on this representation, we simplify the original Fisher criterion in terms of class representation by adding certain constraints on data distribution step by step, and finally approximate it compactly with only several principal component vectors of all the classes. 
Since a set of the principal component vectors of each class span a class subspace, our simplified criterion is defined on the basis of the geometrical relationship between the class subspaces. In this sense, we name this new type of Fisher discriminant analysis based on the simplified criterion geometrical Fisher discriminant analysis (gFDA).

We further introduce the normalization of the projected data on the discriminant space. 
This normalization is in principle required to get 
the best performances out of gFDA and GDS projection, which will be clearly explained through the geometrical structure in Sec.\ref{effectnessNorm}.

The discriminant criterion of gFDA leads to a generalized eigenvalue problem for the matrix product of between-class and within-class matrices.
This formulation makes it difficult to examine the connection of gFDA and GDS projection. Thus, we transform the generalized eigenvalue problem to a simpler regular eigenvalue problem for the linear combination of between-class and within-class matrices.
The linear combination formation leads us to an observation that gFDA is equivalent to GDS projection with a small correction term under a condition of no overlap between class subspaces. As a consequence, we can verify the close connection between FDA and GDS projection via gFDA, as gFDA can be regarded as an approximation of FDA. 

The linear combination formation also enables gFDA to deal with the situation where only a few samples are available. In such a situation, the within-class matrix becomes singular so that FDA cannot in principle be computed. This problem is called the small size sample (SSS) problem of FDA \cite{fda2}. To address the SSS problem, many types of extensions of FDA have been proposed \cite{overviewLDA, newLDA, regFDA, subspaceLDA}.
gFDA is largely different from these conventional methods in that it bypasses the SSS problem by representing the discriminant criterion in a form of linear combination, which can be solved without depending on the number of samples. gFDA can work even with only one sample without any specific modification, unlike most of the above extensions.

Besides, the subspace representation can enhance the robustness of gFDA against the SSS problem. In many applications, a class subspace can be stably generated from even few data; for example, in 3D object recognition, it is well known that a set of the images of a 3D convex object with Lambertian reflectance under various illumination conditions can be represented by a subspace with low dimension (from 3 to 9), which is called illumination subspace \cite{basri, Belhumeur, illuminationSpace}. This means that an illumination subspace of a 3D object like face can be stably and accurately estimated from only a small number (from 3 to 9) of face images under different illuminations. This characteristic of subspace representation works effectively against the SSS problem.


Our main contributions are summarized as follows:
\begin{itemize}
\item We verify that the projection of data onto the generalized difference subspace, GDS projection, works as a discriminant analysis through a mechanism similar to the Fisher discriminant analysis.

\vspace{2mm}
To show the above nature, 
\begin{itemize}
\item We propose a new discriminant analysis, geometrical Fisher discriminant analysis (gFDA), which maximizes a simplified Fisher criterion under a heuristic principle: 
the directions of sample mean vector and first 
principal component vector of a class are nearly the same. 
\item We prove that gFDA is equivalent to GDS projection with a small correction term.
\item We show the close connection of GDS projection and FDA indirectly by regarding gFDA as an intermediate concept between them.
\end{itemize}
\item We demonstrate the effectiveness of gFDA and GDS projection, through extensive comparison experiments with several extensions of FDA on public databases, the Yale face B+ database and the CMU face database, focusing on the small size sample (SSS) problem under few samples.
\end{itemize}

The rest of this paper is organized as follows. Section~\ref{s:gds} and Section~\ref{s:fda} provide preliminary concepts. In Section~\ref{s:gds}, we describe the concept and the definition of the generalized difference subspace (GDS).
In Section~\ref{s:fda}, we overview the fundamentals of FDA with the Fisher criterion. 
In Section~\ref{s:gfda}, we introduce a heuristic principle on the relationship between the first principal component vector and the mean vector of a class. Then, we simplify the Fisher criterion by using the heuristic relationship and construct the geometrical Fisher discriminant analysis (gFDA) with the simplified criterion. 
In Section~\ref{s:mechanismgFDA}, we describe the geometrical mechanism of gFDA and prove that gFDA has dual forms of objective function. 
In Section~\ref{s:connection}, we show the close connection between FDA and GDS projection via gFDA.
In Section~\ref{s:experiments}, we demonstrate the effectiveness of gFDA through evaluation experiments, focusing on the situation of a small sample size. Section~\ref{s:conclusion} concludes the paper.

\section{Generalized difference subspace}\label{s:gds}
In this section, we describe the concept of generalized difference subspace (GDS). As a preliminary to its definition, we describe how to generate a class subspace from the data set for each class. We then define the difference subspace (DS) for two subspaces and extend DS to GDS.
 
\subsection{Generation of class subspace}
The principal component vectors of a class are obtained by applying the principal component analysis (PCA) without data centering to a set of data from the class.  

Given a set of $n_c$ $L$-dimensional data $\{{\bf{x}}^{c}_i\}_{i=1}^{n_c}$ of class $c$ $(c=1,{\dots},C)$, where an image with $w{\times}h$ pixels is regarded as an $L(=w{\times}h)$ dimensional data $\bf{x}$, the principal component vectors $\{{\bs{\phi}}_i^c\}_{i=1}^{N_c}$ of class $c$ are obtained by the following procedure:
\begin{itemize}
\item[1.] An $L{\times}L$ auto-correlation matrix is computed as ${\bf{R}}_c=\frac{1}{n_c}\sum_{i=1}^{n_c}{\bf{x}}^{c}_{i}{{\bf{x}}^{c}_{i}}^{T}$ from $\{{\bf{x}}^{c}_i\}_{i=1}^{n_c}$.
\item[2.] The principal component vectors $\{{\bs{\phi}}_i^c\}_{i=1}^{N_c}$ of class $c$ are obtained as the unit eigenvectors corresponding to the $N_c$ largest eigenvalues of \({\bf{R}}_{c}\). If we use all the eigenvalues, we obtain the spectral decomposition of the matrix ${\bf{R}}_c$. 
\end{itemize}
Throughout the whole paper, the principal component vectors of a class are used as the orthonormal basis vector of the corresponding class subspace. In the following, we will interchangeably use the terms of principal component vector and orthonormal basis vector as of the same meaning.

\subsection{Geometrical definition of DS}\label{defDS1}
\begin{figure}[bt]
  \begin{center}
    \includegraphics[width=70mm]{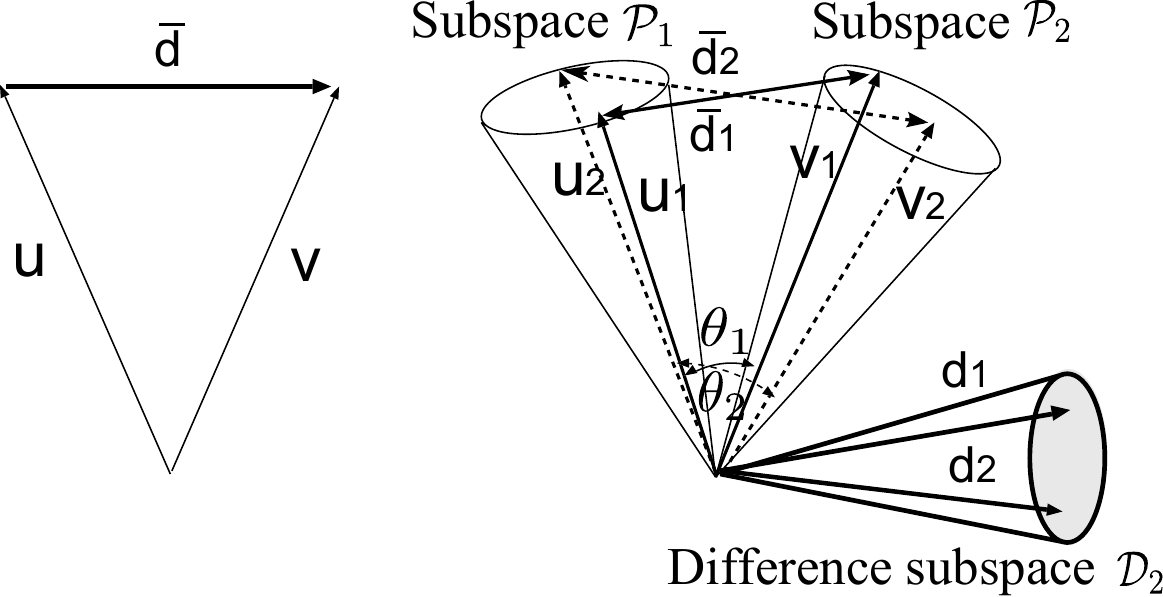}\\
	(a)\hspace{40mm}(b)\\
    \caption{Basis concept of difference subspace: (a) difference vector, (b) canonical angles, vectors and difference subspace.}
    \label{dspace}
  \end{center}
\end{figure}

The $\it difference~subspace$ (DS) is a natural extension of a difference vector $\bar{\bf{d}}$ between two vectors $\bf{u}$ and $\bf{v}$ as shown in Figs.\ref{dspace}a and \ref{dspace}b \cite{gds}. 

We formulate the \(\it difference~subspace\) between \(M\)-dimensional subspace ${\cal{P}}_1$ and \(N\)-dimensional subspace ${\cal{P}}_2$ in $L$-dimensional vector space. In the case that there is no overlap between these subspaces, $N$ canonical angles $\{\theta_{i}\}_{i=1}^N$ (for~convenience \(N \leq M)\) can be obtained between them \cite{cangle1,cangle2}. 
Let \(\bar{{\bf d}}_{i}\) be the difference vector, \({\bf{v}}_{i}-{\bf{u}}_{i}\), between canonical vectors \({\bf{u}}_{i} \in {\cal{P}}_1\) and \({\bf v}_{i} \in {\cal{P}}_2\), which form the \(\it i\)th canonical angle \(\theta_{i}\). All $\bar{\bf d}_i$ are orthogonal to each other. Thus, after normalizing the length of each difference vector \({\bar{\bf d}}_{i}\) to 1, we regard the normalized difference vectors ${\bf{d}}_i=\frac{{\bf{v}}_i -{\bf{u}}_i}{||{\bf{v}}_i -{\bf{u}}_i||}$ as the orthonormal basis vectors of the \(\it{difference~subspace}\) \({\cal{D}}_{2}\). Thus, ${\cal{D}}_{2}$ is defined as $<{{\bf d}}_{1}, {{\bf d}}_{2}, \cdots, {{\bf d}}_{N}>$.

\subsection{Analytical definition of DS}
The \(\it difference~subspace\) defined geometrically in Sec.\ref{defDS1} can also be analytically defined by  using the orthogonal projection matrices of two class subspaces \cite{gds}.

\vspace{2mm}\noindent
{\bf Theorem.} {\it The $i$-th basis vector ${\bf{d}}_i$ of the difference subspace ${\cal{D}}_2$ is equal to the normalized eigenvector ${\bf{x}}_i$ of ${\bf{P}}_1 + {\bf{P}}_2$ that corresponds to the $i$-th smallest eigenvalue smaller than 1, where ${\bf{P}}_1$ and ${\bf{P}}_2 \in {\bf{R}}^{L{\times}L}$ are the orthogonal projection matrices of the two class subspaces, defined by $\sum_{i=1}^M {\bs{\phi}}_i^1 {{\bs{\phi}}_i^{1}}^{T}$ and $\sum_{i=1}^N {\bs{\phi}}_i^2 {{\bs{\phi}}_i^{2}}^{T}$, respectively.}

\begin{figure}[bt]
  \begin{center}
    \includegraphics[width=80mm]{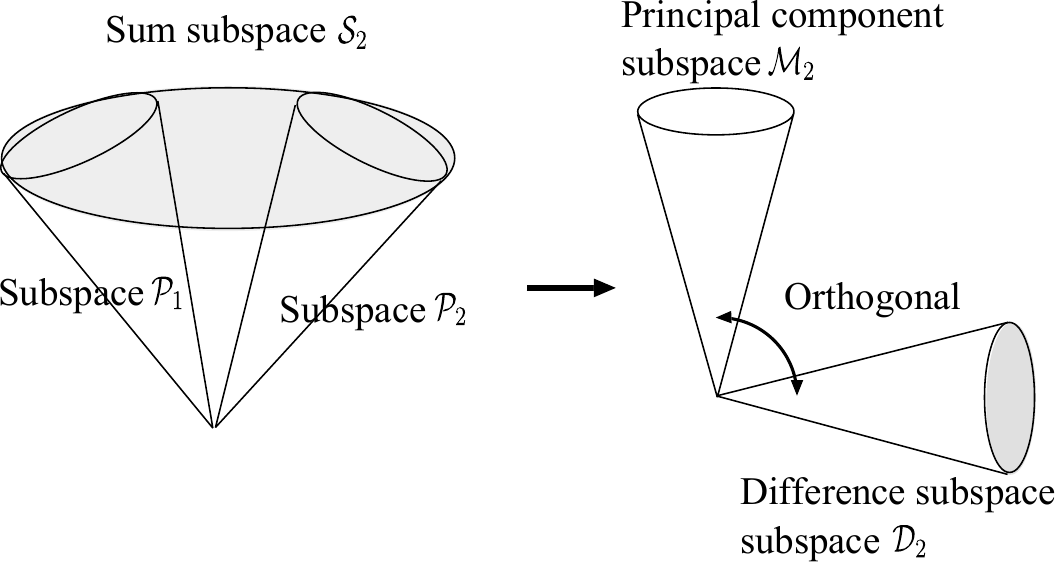}\\
    \caption{Direct sum decomposition of sum subspace \({\cal{S}}_2\) into principal component subspace \({\cal{M}}_2\) and difference subspace \({\cal{D}}_2\).}
    \label{sumspace}
  \end{center}
\end{figure}

\begin{itemize}
\item $N$ eigenvectors of matrix \({\bf{P}}_1 + {\bf{P}}_2\) corresponding
to eigenvalues smaller than 1 span the \(\it{difference~subspace}\)~
\({\cal{D}}_2\).
\item $N$ eigenvectors of matrix \({\bf{P}}_1 + {\bf{P}}_2\) corresponding
to eigenvalues larger than 1 span the \(\it{principal~component~subspace}\)~\({\cal{M}}_2\).
\end{itemize}

The relations lead to the conclusion that the sum subspace~\({\cal{S}}_2\) of ${\cal{P}}_1$ and ${\cal{P}}_2$, spanned by all the eigenvectors of matrix \({\bf{P}}_1+{\bf{P}}_2\), is represented by the orthogonal direct sum of the principal~component~subspace~\({\cal{M}}_{2}\) and the \(\it{difference~subspace}\)~\({\cal{D}}_2\) as ${\cal{S}}_2={\cal{M}}_2{\bigoplus}{\cal{D}}_2$. Fig.\ref{sumspace} shows the conceptual diagram of this direct sum decomposition. 
This means that the \(\it{difference~subspace}\) \({\cal{D}}_2\) can be defined as the subspace that is produced by removing the principal~component~subspace~\({\cal{M}}_{2}\) from the sum subspace~\({\cal{S}}_2\). Hence, the \(\it{difference~subspace}\) can be regarded as the subspace that does not include the principal component of the two subspaces, that is, it contains only the difference component between them.

\subsection{Definition of GDS}
\begin{figure}[bt]
  \begin{center}
    \includegraphics[width=55mm]{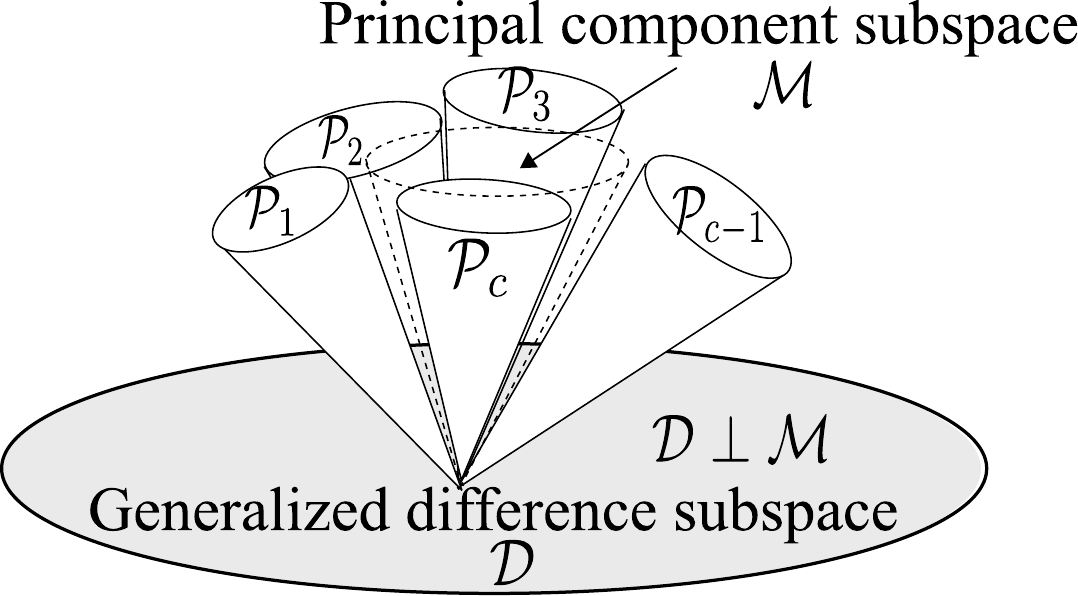}
    \caption{Conceptual diagram of generalized difference subspace for \(c\) subspaces.}
    \label{cspace}
  \end{center}
\end{figure}

To deal with the difference between two or more subspaces, the concept of the {\it difference~subspace} was generalized under the analytical definition \cite{gds}. Fig.\ref{cspace} shows the conceptual diagram of the {\it generalized~difference~subspace} (GDS) \({\cal{D}}\) for $C$ subspaces. 

Given \(C({\geq}2)\) \(N_c\)-dimensional subspaces $\{{\cal{P}}_c\}_{c=1}^C$ in $L$-dimensional vector space, a {\it generalized~ difference~subspace} \({\cal{D}}\) can be defined as such a  subspace that is produced by removing the principal component subspace \({\cal{M}}\), of all the subspaces, from the sum subspace \({\cal{S}}\) of ${\{{\cal{P}}_c}\}_{c=1}^C$.
Thus, the {\it generalized~difference~subspace} \({\cal{D}}\) is spanned by \(N_d\) eigenvectors, ${\{{\bf{d}}_{i}\}}_{i=1}^{N_d}$ corresponding to the \(N_d\) smallest eigenvalues, of the following sum matrix ${\bf{G}}$:
\begin{eqnarray}
{\bf{G}}=\sum_{c=1}^{C}{\bf{P}}_c=\sum_{c=1}^{C}\sum_{i=1}^{N_c} {\bs{\phi}}_i^c {{\bs{\phi}}_i^{c}}^{T}~,  
\label{defofG}
\end{eqnarray}
where ${\bf{P}}_c \in {\bf{R}}^{L{\times}L}$ denotes the orthogonal projection matrix of the class $c$ subspace.

The generalized difference subspace \({\cal{D}}\) contains only the essential component for discriminating all the classes, since it is the orthogonal complement of the principal component subspace \({\cal{M}}\) that represents the common information of all the class subspaces. 

\section{Fisher discriminant analysis}\label{s:fda}
Fisher discriminant analysis (FDA) is a method for obtaining a discriminant space $\cal{F}$, which can distinguish multiple classes effectively \cite{fda1,fda2}. Such a discriminant space can be found out by maximizing the Fisher criterion of the projected data on the discriminant space $\cal{F}$. 

The Fisher criterion consists of within-class covariance matrix and between-class covariance matrix. Given $C$ classes, each of which contains the data set $\{{\bf{x}}_i^c\}_{i=1}^{n_c}$ ($c=1,\dots,C$), the within-class covariance matrix $\SW \in {\bf{R}}^{L{\times}L}$ is defined as
\begin{eqnarray}
\SW = \frac{1}{n} \sum_{c=1}^C \sum_{i=1}^{n_c}  ({\bf{x}}_i^c-{\bf{m}}_c)  {({\bf{x}}_i^c-{\bf{m}}_c) }^T, \label{sw1}
\end{eqnarray}
where $n=\sum_{c=1}^C n_c$ and ${\bf{m}}_c$ indicates the mean vector of class $c$.
The between-class covariance matrix ${\SB} \in {\bf{R}}^{L{\times}L}$ is defined by the following equation:
\begin{eqnarray}
\SB =  \frac{1}{n}\sum_{c=1}^C  {n_c}({\bf{m}}_c -{\bf{m}}) {({\bf{m}}_c -{\bf{m}})}^T  \label{sb0},
\end{eqnarray}
where $\bf{m}$ indicates the mean vector over all the classes; ${\SB}$ can also be represented by 
\begin{eqnarray}
\SB =  \frac{1}{n^2} \sum_{i=1}^{C-1}  \sum_{j=i+1}^C  {n_i n_j} ({\bf{m}}_i-{\bf{m}}_j) {({\bf{m}}_i -{\bf{m}}_j)}^T  \label{between2}.
\end{eqnarray}

The Fisher criterion $f({\bf{d}})$ of the data projected on a 1-dimensional subspace spanned by vector ${\bf{d}}$ is defined as
\begin{eqnarray}
f({\bf{d}}) &=& \frac{{\bf{d}}^{T} {\SB} {\bf{d}}}{{\bf{d}}^{T} {\SW} {\bf{d}}}, \label{defFR}
\end{eqnarray}
where the vector $\bf{d}$ that maximizes function $f$ can be obtained by solving the generalized eigenvalue problem
\begin{eqnarray}
{\SB}{\bf{d}}= \lambda {\SW}{\bf{d}}.
\end{eqnarray}
Discriminant space $\cal{F}$ is spanned by $C-1$ eigenvectors, $\{{\bf{d}}_i\}_{i=1}^{C-1}$, corresponding to the $C-1$ largest eigenvalues of the above eigenvalue problem.

\section{Geometrical fisher discriminant analysis}\label{s:gfda}
In this section, we first approximate the Fisher criterion, $f({\bf{d}})$ in Eq.(\ref{defFR}), based on a heuristic principle. We then simplify it by adding constraints on data distribution in incremental steps. Finally, we construct the proposed gFDA by maximizing the simplified Fisher criterion.


\subsection{Equivalence between the class mean and first principal component vector}
We introduce a heuristic principle on the equivalence between the first principal component vector ${\bs{\phi}}_1^c$ and the mean vector ${\bf{m}}_c$ of class $c$.

\vspace{2mm}
\noindent {\bf{Heuristic principle}}: For each class subspace, the first principal component vector ${\bs{\phi}}_1^c$ and the mean vector ${\bf{m}}_c$ can be in a very close correspondence with each other in terms of their directions, under the condition that ${||{\bf{m}}_c||}^2 \gg  \sigma_{max}^2$, where $\sigma_{max}^2$ is the maximum variance of the class distribution among all the dimensions.
\vspace{2mm}

\begin{figure}[bt]
  \begin{center}
    \includegraphics[width=85mm]{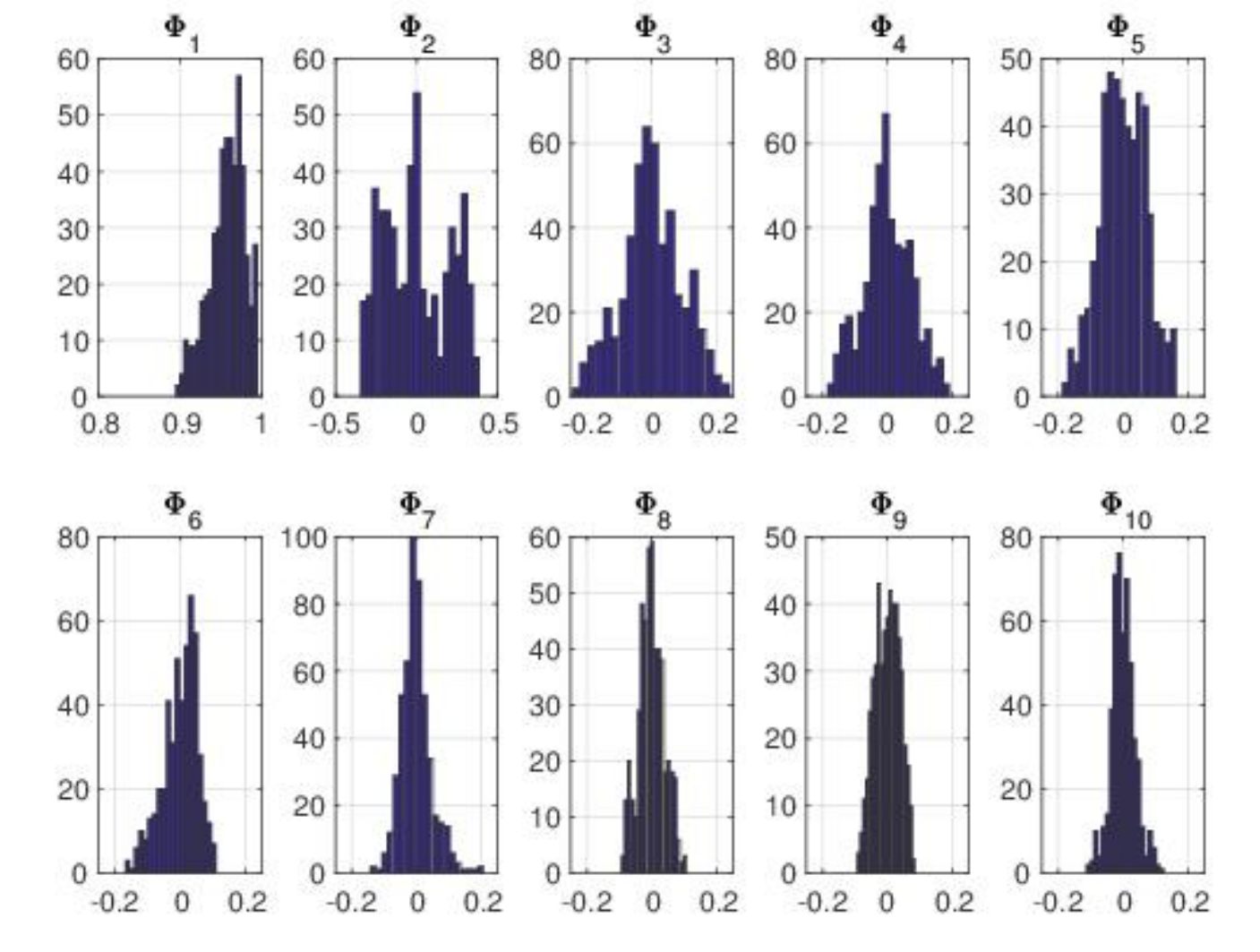}
\caption{Histograms of data distributions projected on ten different principal component vectors.}    \label{visualPCA}
  \end{center}
\end{figure}

Under this heuristic principle, the first and the remaining principal component vectors
have different characteristics; the projected data on the first principal component vector should have a positive mean value, while those on the remaining vectors should have zero mean value. 
Fig.\ref{visualPCA} shows an example of the histograms of the projections on each principal component vector in the case of front face images. We can observe that the means of the projections onto the principal component vectors are all nearly zero except the first one, which supports that the heuristic principle should be valid in real data.

The mechanism that the condition of ${||{\bf{m}}_c||}^2 \gg  \sigma_{max}^2$ yields our heuristic principle can be considered as follows.
%
Given a data set, ${ \{{\bf{x}}_i^c \}}_{i=1}^{n_c}$ of class $c$, the autocorrelation matrix ${\bf{R}}_c$ and the covariance matrix ${\bf{C}}_c$ are defined as
\begin{eqnarray}%
{\bf{R}}_c&=& \frac{1}{n_c}\sum_{i=1}^{n_c} {\bf{x}}_i^c {{\bf{x}}_i^c}^T,\\
{\bf{C}}_c&=& \frac{1}{n_c}\sum_{i=1}^{n_c} ({\bf{x}}_i^c-{\bf{m}}_c) ({\bf{x}}_i^c-{\bf{m}}_c)^T,
\end{eqnarray}
where ${\bf{m}}_c$ indicates the mean vector of the class. There is the following relationship between the two matrices and the mean vector:
\begin{eqnarray}%
{\bf{R}}_c={\bf{C}}_c+{\bf{m}}_c {\bf{m}}_c^T.
\end{eqnarray}
The condition of ${\|{\bf{m}}_c\|}^2 \gg  {\sigma}_{max}^2$ ensures ${\|{\bf{m}}_c {{\bf{m}}}_c^T\|}_F^2$ $\gg$ ${\|{\bf{C}}_c\|}_F^2$, thus ${\bf{m}}_c$ is dominant in constructing ${\bf{R}}_c$. Therefore, the direction of the first principal component vector ${\bs{\phi}}_1$ corresponding to the largest eigenvalue of ${\bf{R}}_c$ almost coincides with that of vector ${\bf{m}}_c$.

The heuristic principle can work with high degree of coincidence even under a loose condition. 
According to a simulation using randomly generated Gaussian distributions in vector spaces with various dimensions, the directions of them coincided with high correlation of over 0.998 even under the condition that $\frac{{\|{\bf{m}}_c\|}}{\sigma_{max}} \geq 2$.
In many tasks, for example, object image classification, the condition of $\frac{{\|{\bf{m}}_c\|}}{\sigma_{max}} \geq 2$ can hold in most cases.
Furthermore, we will experimentally confirm the validity of the heuristic principle on face data in Section~\ref{s:experiments}. 

\subsection{Simplification of the Fisher criterion}
The within-class covariance matrix $\SW \in {\bf{R}}^{L{\times}L}$ defined in Eq.(\ref{sw1}) can be rewritten by the autocorrelation matrix ${\bf{R}}_c$ and the mean vectors ${\bf{m}}_c$ of the $c$th class as follows:
\begin{eqnarray}
\SW = \frac{1}{n}\sum_{c=1}^C (n_c {\bf{R}}_c - n_c {\bf{m}}_c {{\bf{m}}_c}^T),
\label{sw0}
\end{eqnarray}
where $n=\sum_{c=1}^C n_c$. By using the spectral decomposition of ${\bf{R}}_c$, $\SW$ can be rewritten as
\begin{eqnarray}
\SW = \frac{1}{n}\sum_{c=1}^C n_c \Bigl((\sum_{i=1}^{{d}_{c}^{all}}  \lambda_i^c  {\bs{\phi}}_i^c  {{\bs{\phi}}_i^c}^T ) - {\bf{m}}_c {{\bf{m}}_c}^T \Bigr),
\end{eqnarray}
where $d_c^{all}= {\rm min}(n_c, L)$, and $\lambda_i^c$ and ${\bs{\phi}}_i^c$ indicate the $i$th eigenvalue of the autocorrelation matrix  ${\bf{R}}_c$ of the class $c$ and its corresponding eigenvector, respectively.

Furthermore, by using the heuristic principle, ${\bf{m}}_c \approx {\bar{m}}_c {\bs{\phi}}_1^c$ where ${\bar{m}}_c={\|{\bf{m}}_c\|}_2$, and assuming that 
$n_c$ = ${\bar{n}}$ for all the classes, we replace ${\SW}$ with ${\SW}_1$: 
\begin{eqnarray}
{\SW}_1 &=& \frac{\bar{n}}{n}\sum_{c=1}^C \Bigl((\sum_{i=1}^{{d}_c^{all}}  \lambda_i^c  {\bs{\phi}}_i^c  {{\bs{\phi}}_i^c}^T) - {\bar{m}}_c^2{\bs{\phi}}_1^c {{\bs{\phi}}_1^c}^T \Bigr), \\
&=& \frac{\bar{n}}{n}\sum_{c=1}^C \sum_{i=1}^{{d}_c^{all}}  {\sigma_i^c}^2  {\bs{\phi}}_i^c  {{\bs{\phi}}_i^c}^T,
\end{eqnarray}
where ${\sigma_i^c}^2$ represents the variance of the data projected on the $i$th principal component vector ${\bs{\phi}}_i^c$, ${\sigma_1^c}^2=\lambda_1^c- {\bar{m}}_c^2$ and ${\sigma_i^c}^2 = \lambda_i^c\ (i \geq 2)$.

With the heuristic principle, the between-class covariance $\SB$ can be represented with ${\bs{\phi}}_1^c$ as follows:
\begin{equation}
{\SB}_1 =  \frac{{\bar{n}}^2}{n^2} \sum_{i=1}^{C-1}  \sum_{j=i+1}^C   
({\bar{m}}_i {\bs{\phi}}_1^i-{\bar{m}}_j {\bs{\phi}}_1^j)
{({\bar{m}}_i {\bs{\phi}}_1^i-{\bar{m}}_j {\bs{\phi}}_1^j)}^T. \label{sb1}
\end{equation}

We refer to an FDA based on the Fisher criterion of \(\frac{{\bf{d}}^T{\SB}_1{\bf{d}}}{{\bf{d}}^T{\SW}_1{\bf{d}}}\) as approximated FDA (aFDA), which is more or less equivalent to the original FDA. 
We simplify the representation of ${\SB}_1$ and ${\SW}_1$ in the following two steps.

\vspace{2mm}
\noindent {\bf Simplification-I}:
First, we use only $N_c$ eigenvectors corresponding to the larger eigenvalues than a specified threshold and discard the eigenvectors corresponding to smaller eigenvalues:
\begin{eqnarray}
{\SW}_2 = \frac{\bar{n}}{n}\sum_{c=1}^C \sum_{i=1}^{N_c}  {\sigma_i^c}^2  {\bs{\phi}}_i^c  {{\bs{\phi}}_i^c}^T,
\end{eqnarray}
Moreover, assuming that the norms ${\bar{m}}_i$ of the mean vectors of all the classes are equal to ${\bar{m}}$,
we can simplify ${\SB}_1$ to ${\SB}_2$ using $n=C{\bar{n}}$ as follows: 
\begin{eqnarray}
{\SB}_2 &=&  \frac{\bar{m}^2 {\bar{n}}^2}{C^2{\bar{n}}^2} \sum_{i=1}^{C-1}  \sum_{j=i+1}^C   
({\bs{\phi}}_1^i-{\bs{\phi}}_1^j) ({{\bs{\phi}}_1^i-{\bs{\phi}}_1^j)}^T,
\end{eqnarray}
\begin{eqnarray}
 &=&  \frac{\bar{m}^2}{C^2} \sum_{i=1}^{C-1}  \sum_{j=i+1}^C   
({\bs{\phi}}_1^i-{\bs{\phi}}_1^j) ({{\bs{\phi}}_1^i-{\bs{\phi}}_1^j)}^T \label{sb2}.\end{eqnarray}

We refer to an FDA based on the simplified Fisher criterion of \(\frac{{\bf{d}}^T{\SB}_2{\bf{d}}}{{\bf{d}}^T{\SW}_2{\bf{d}}}\) as simplified FDA (sFDA). 
\begin{figure}[bt]
  \begin{center}
\includegraphics[width=90mm]{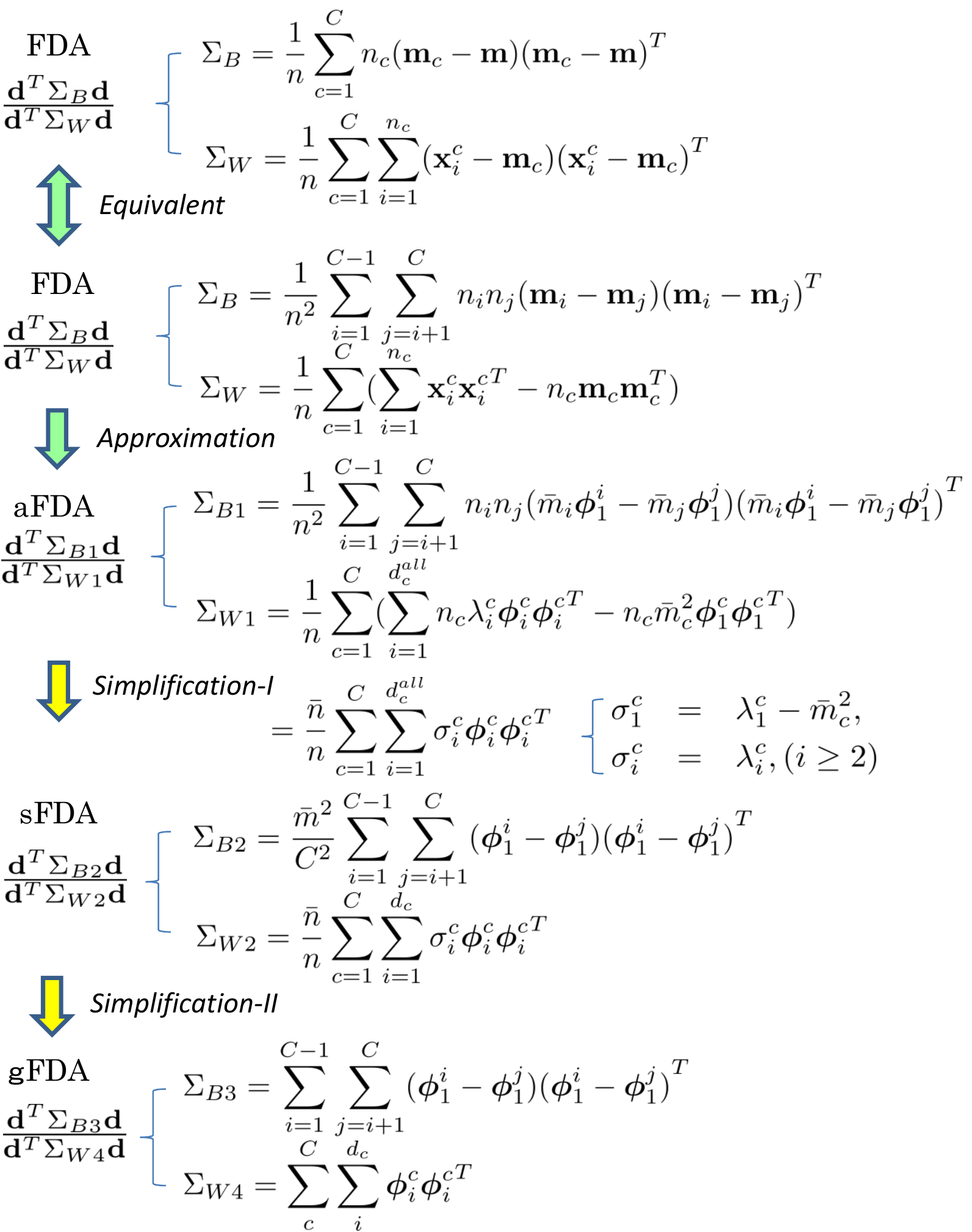}
\caption{The simplification process from FDA toward gFDA. the simplification process from Eq.(\ref{sw1}) to Eq.(\ref{sw3}) is summarized.}
\label{fromsFDAtogFDA}
  \end{center}
\end{figure}

\vspace{2mm}
\noindent {\bf Simplification-II}:
Next, we assume that all the values of $\{{\sigma_i^c}^2\}$ are equally ${\bar{\sigma}}^2$.
This assumption can enhance sFDA's robustness 
against few training data as will be shown later, although one may feel it extreme.
Under this assumption, we further simplify ${\SW}_2$ to ${\SW}_{3}$ using $n=C{\bar{n}}$ as
\begin{eqnarray}
{\SW}_3 = \frac{{\bar{\sigma}}^2}{C} \sum_{c=1}^C \sum_{i=1}^{N_c}  {\bs{\phi}}_i^c  {{\bs{\phi}}_i^c}^T.
\end{eqnarray}
It is possible to define several types of Fisher-like criteria as combinations of the above simplified matrices. In this paper, we are interested in the simplest criterion defined by ${\SB}_2$ and ${\SW}_3$ and consider the following objective function $f_1$:
\begin{eqnarray}
{f_1}({\bf{d}}) &=& \frac{{\bf{d}}^{T} {\SB}_2 {\bf{d}}}{{\bf{d}}^{T} {\SW}_3 {\bf{d}}}, \label{f1:0}\\
&=&  \frac{{\bar{m}}^2}{{{\bar{\sigma}}^2 C}}  \frac{{\bf{d}}^{T} {\SB}_3 {\bf{d}}}{{\bf{d}}^{T} {\SW}_4 {\bf{d}}}, \label{f1:1}\\
{\rm where} \nonumber\\
{\SB}_3 &=& \sum_{i=1}^{C-1}  \sum_{j=i+1}^{C} ({\bs{\phi}}_1^i-{\bs{\phi}}_1^j) ({{\bs{\phi}}_1^i-{\bs{\phi}}_1^j)}^T,\\
{\SW}_4 &=& \sum_{c=1}^C \sum_{i=1}^{N_c}  {\bs{\phi}}_i^c  {{\bs{\phi}}_i^c}^T. \label{sw3}
\end{eqnarray}

Since the term of $\frac{{\bar{m}}^2}{{{\bar{\sigma}}^2 C}}$ is constant, we can ignore it and obtain vector $\bf{d}$ by maximizing the following objective function $f_g$ instead of $f_1$:
\begin{eqnarray}
{f_g}({\bf{d}}) = \frac{{\bf{d}}^{T} {\SB}_3 {\bf{d}}}{{\bf{d}}^{T} {\SW}_4 {\bf{d}}}. \label{fg} \label{ourCriteria}
\end{eqnarray}
The process of the set of simplifications is summarized in Fig.\ref{fromsFDAtogFDA}. We define an FDA based on the above simplified Fisher criterion as geometrical FDA (gFDA). Finally, vectors ${\bf{d}}$ that maximizes $f_g$ are obtained by solving the following generalized eigenvalue problem:
\begin{eqnarray}
{\SB}_3 {\bf{d}}=\lambda {\SW}_4 {\bf{d}}. \label{fg2}
\end{eqnarray}

\begin{figure}[bt]
\begin{center}
\subfloat[2 classes]{
\begin{tabular}{cc}
\includegraphics[width=40mm]{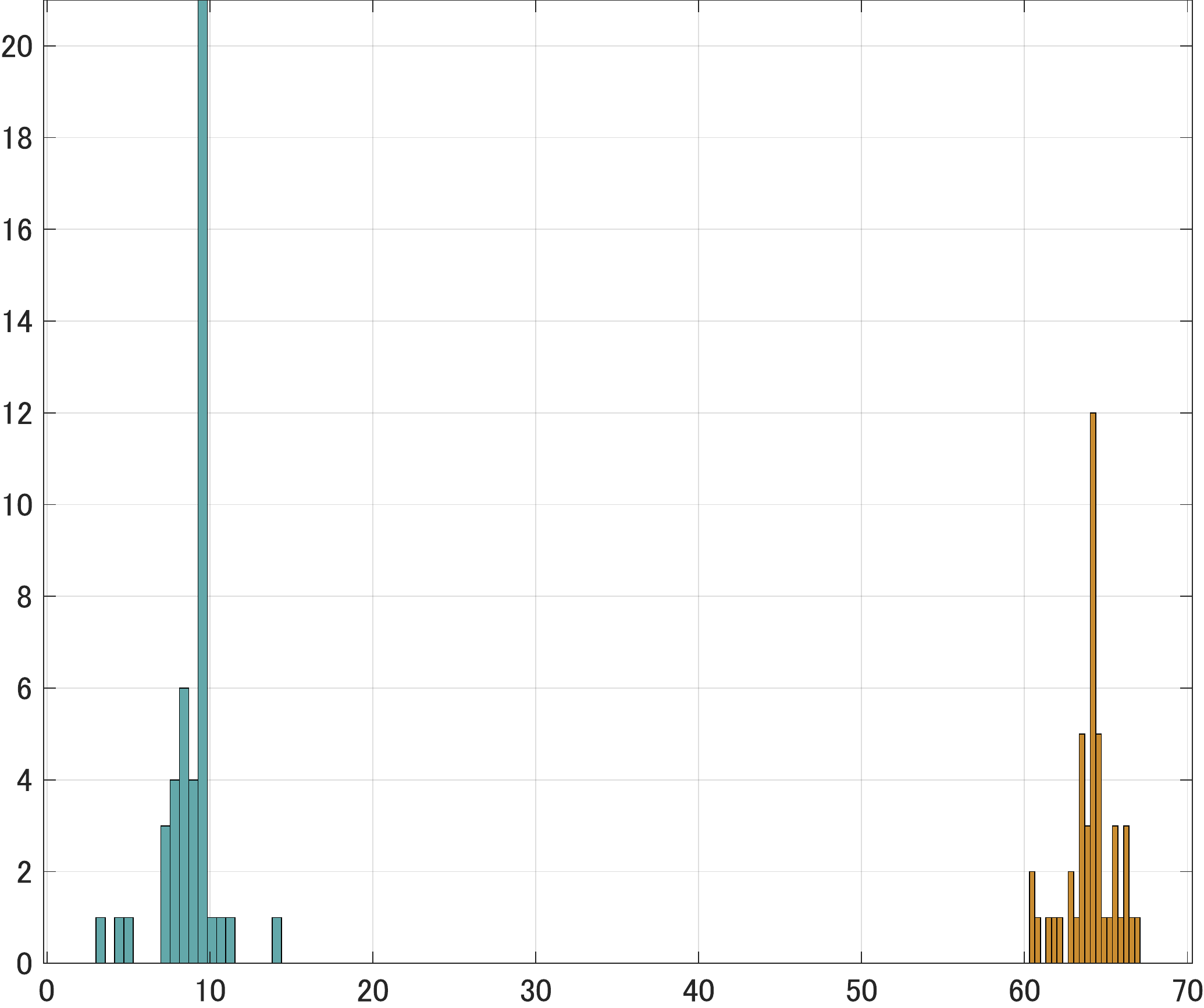}&
\includegraphics[width=38mm]{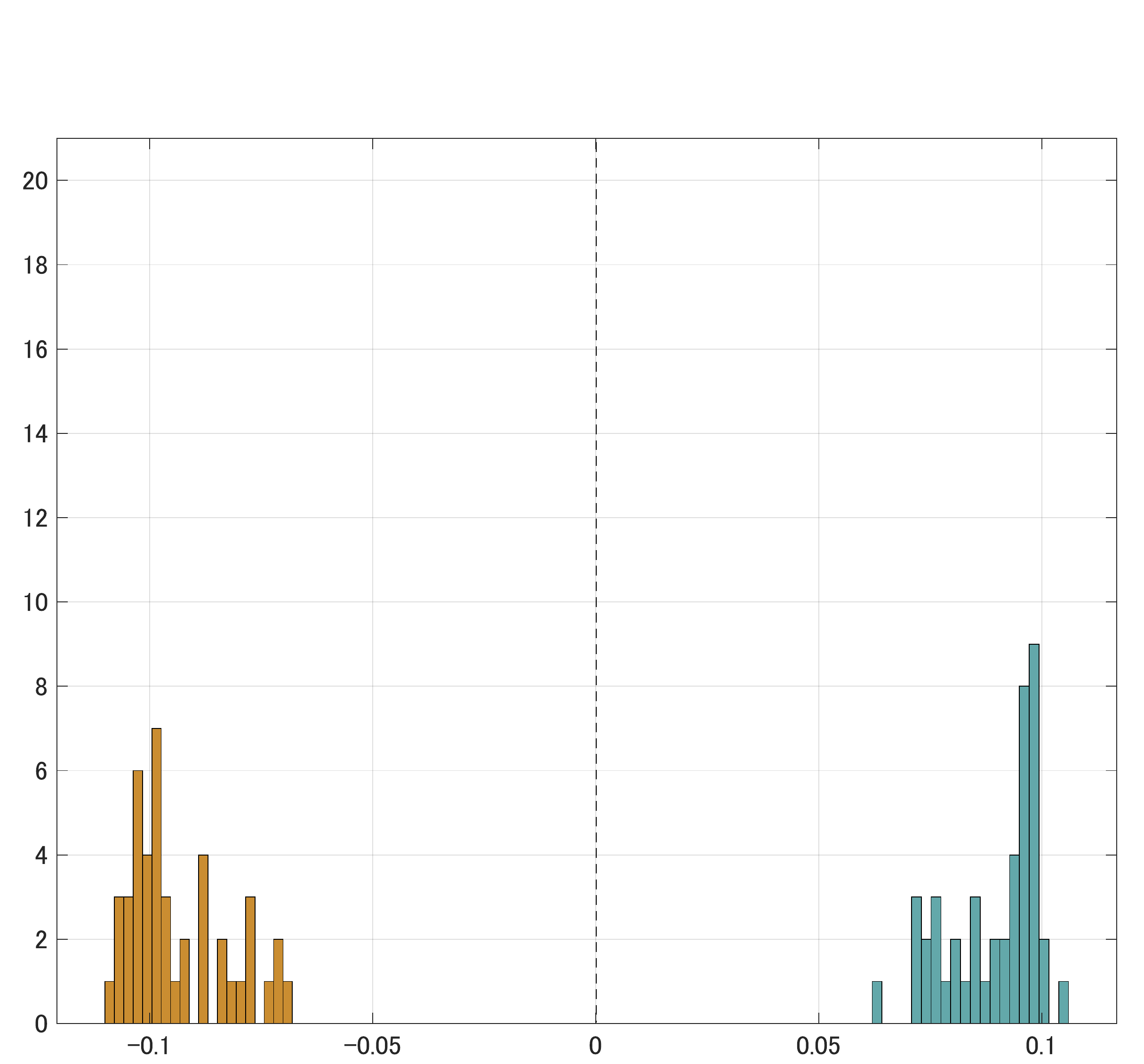}
\end{tabular}
}\\
\subfloat[3 classes]{
\begin{tabular}{cc}
\includegraphics[width=42mm]{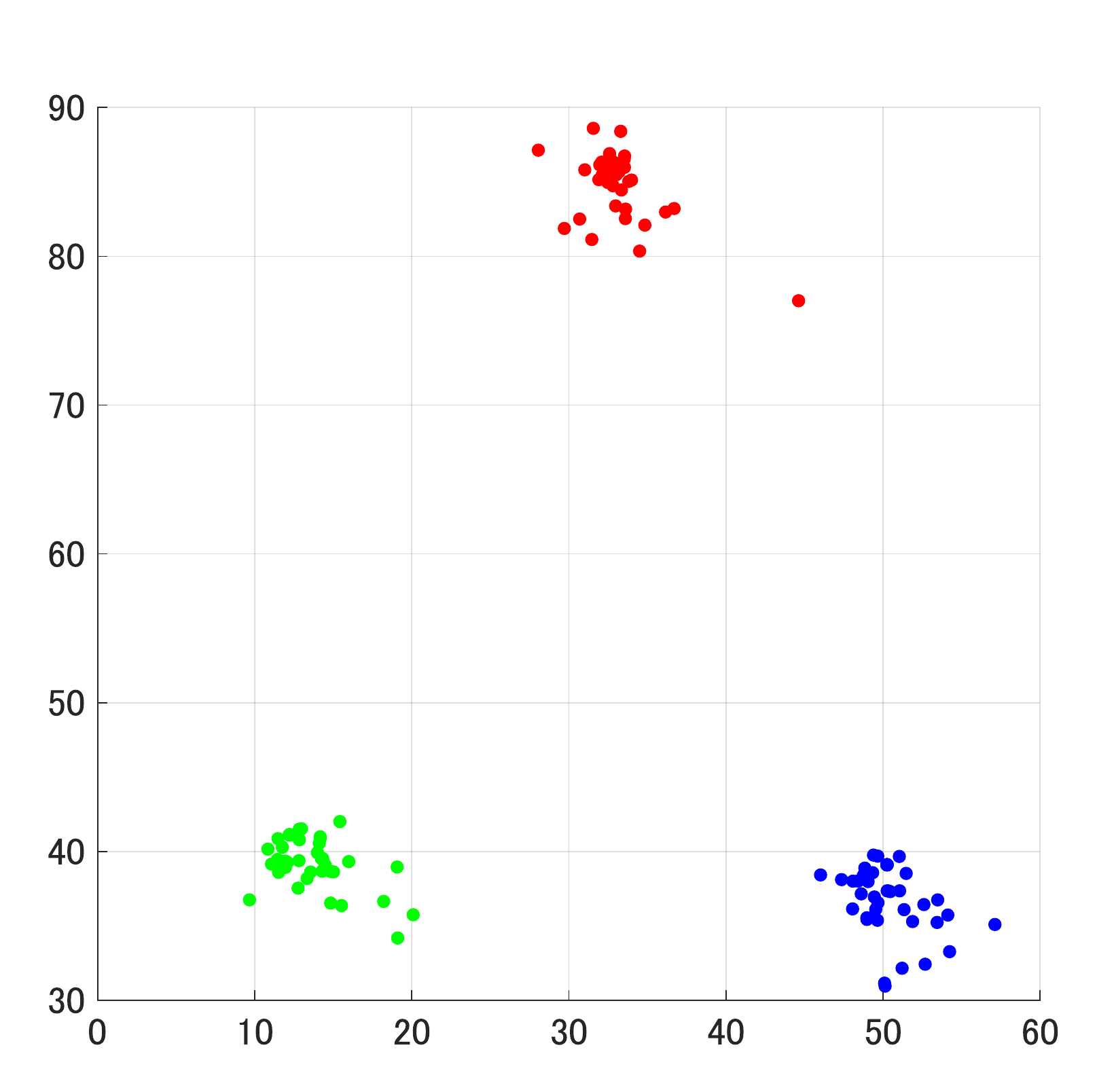}&
\includegraphics[width=39mm]{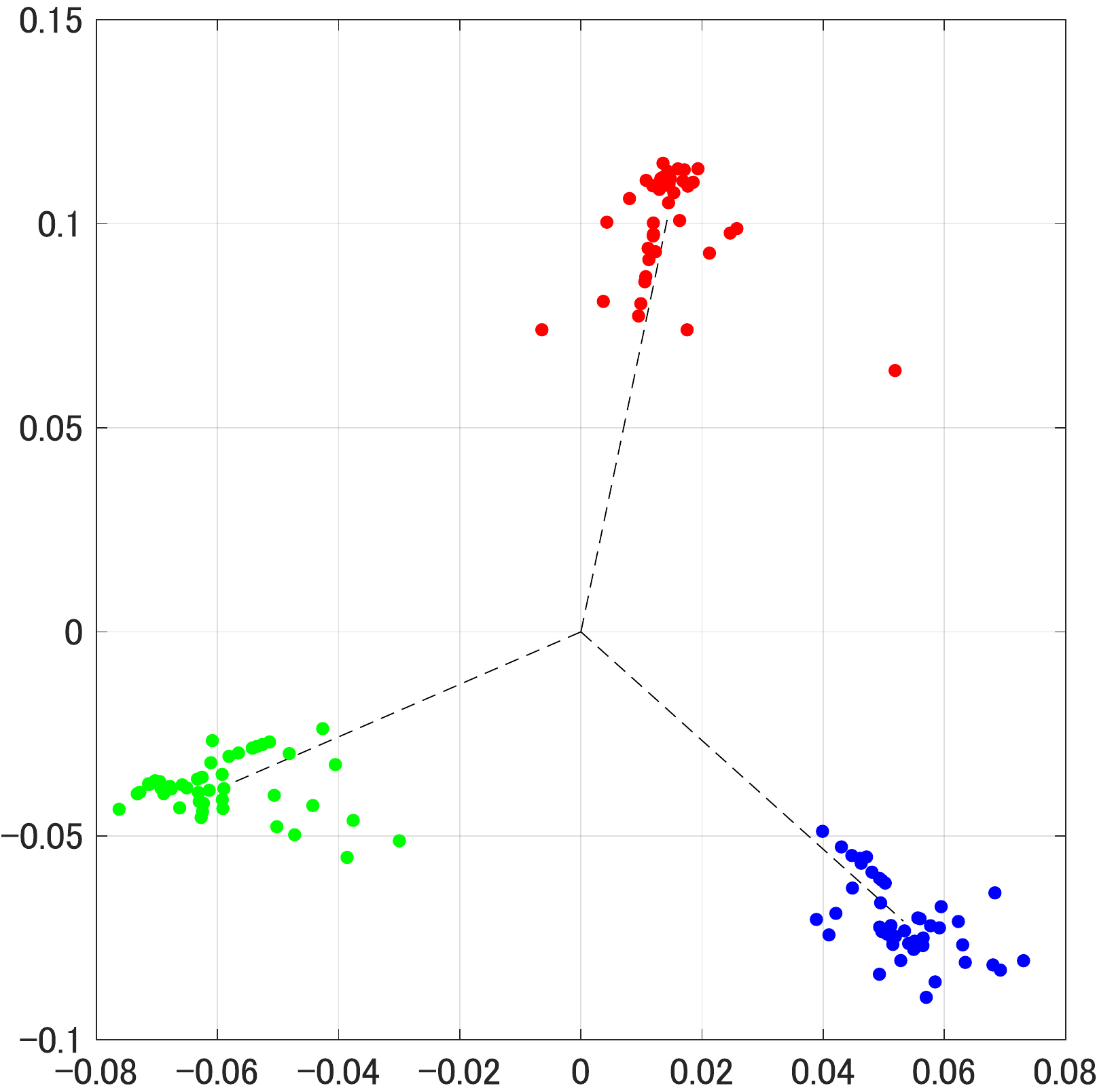}
\end{tabular}
}\\
\subfloat[4 classes]{
\begin{tabular}{cc}
\includegraphics[width=41mm]{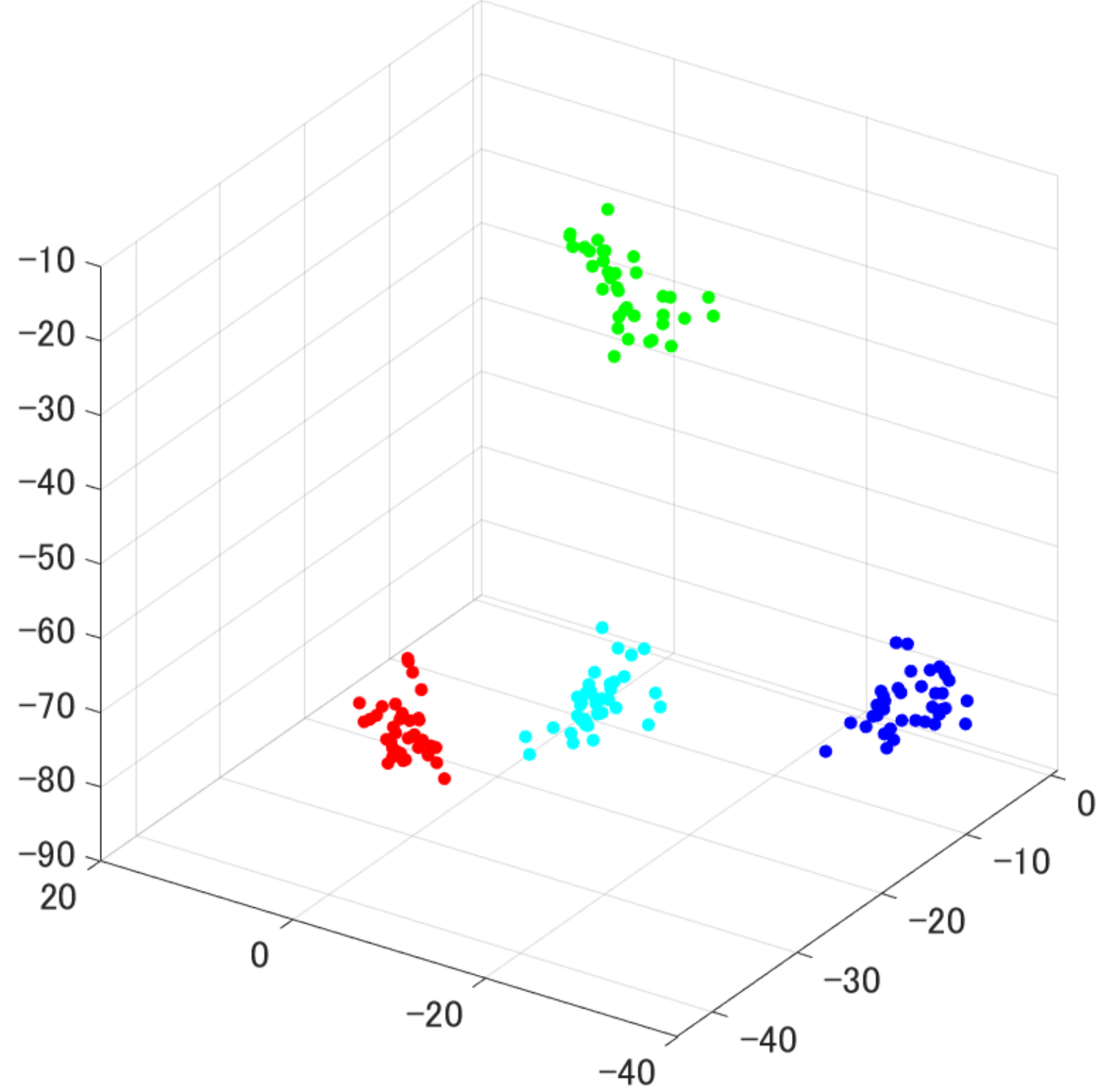}&
\includegraphics[width=43mm]{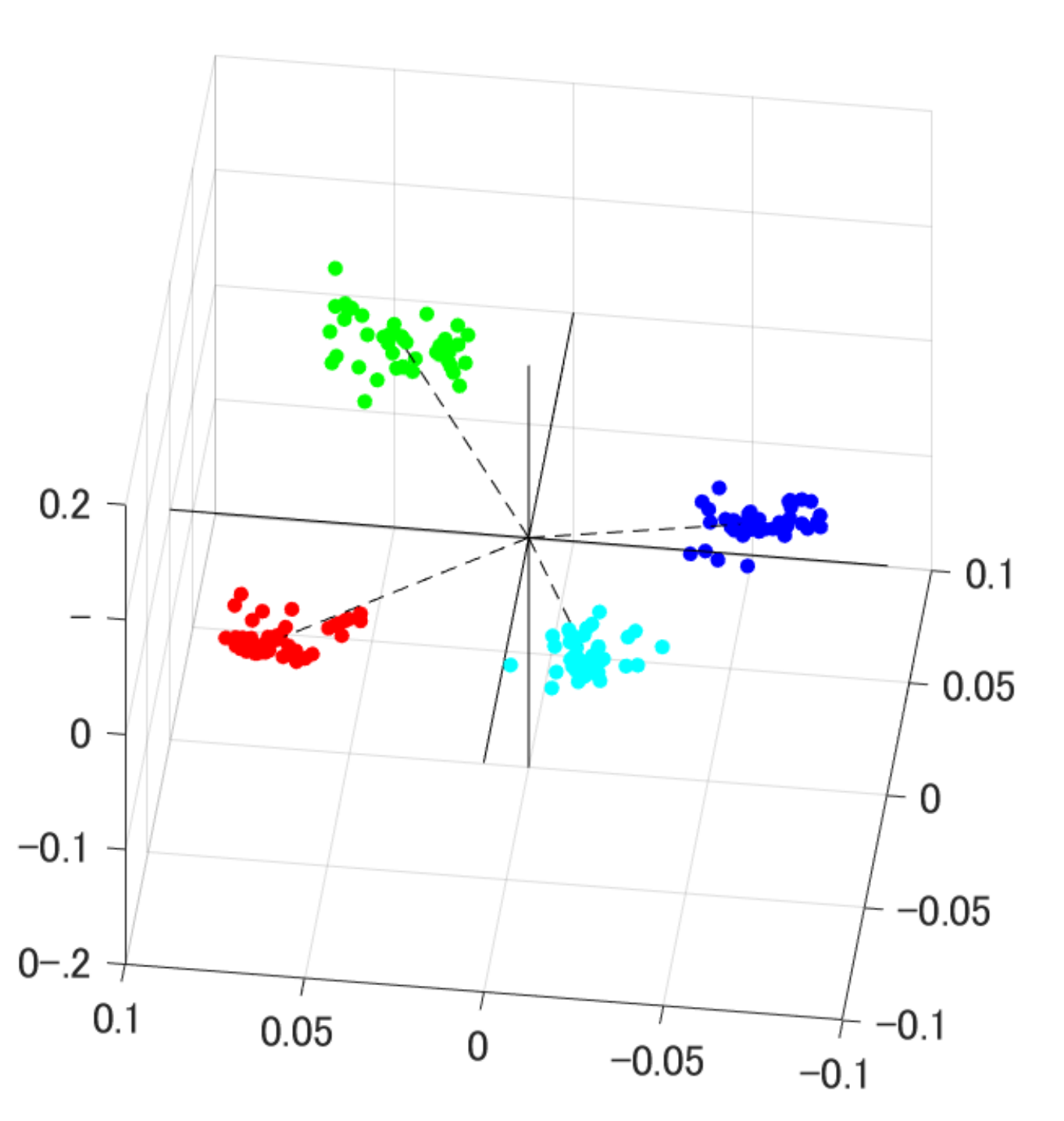}
\end{tabular}
}\\
\caption{The projections of face classes from the Yale face database by FDA (left) and gFDA (right), where a row represents the case of 2, 3 or 4 classes.} 
\label{comp_fda_gfda}
\end{center}
\end{figure}

\subsection{Small sample size problem}\label{maximization}
In many practical applications, the dimension $L$ of data is much larger than the total number of data, $n$. In such a case, Eq.(\ref{fg2}) cannot be solved since ${{\SW}}_4$ is singular. This issue is called the small sample size (SSS) problem \cite{fda2}, which has been well known as a critical limitation of FDA.

To overcome the SSS problem, various types of extensions of FDA have been proposed \cite{overviewLDA}.
There are two typical solutions widely used due to their simple implementation. One is to use PCA to reduce the dimension before applying FDA \cite{fda2}. The other is to add a regularization term to matrix ${\SW}$ \cite{regFDA} as follows:
\begin{eqnarray}
f({\bf{d}}) = 
\frac{{\bf{d}}^{T} {\SB} {\bf{d}}}
{{\bf{d}}^{T} ({\SW} + \delta {\bf{E}}){\bf{d}}}, 
\label{f2}
\end{eqnarray}
where $\delta$ is a parameter that controls the strength of the regularization and $\bf{E}$ is the identity matrix.

In addition to the above simple methods, nullLDA \cite{newLDA} is also often used to address the SSS problem. In this method, all the data are first projected onto the null space spanned by the within-class scatter matrix, and then a between-class scatter matrix is calculated from the projections. Finally, a discriminant space is obtained by solving the eigenvalue problem of the between-class scatter matrix. Many extensions of FDA based on similar ideas have been proposed to circumvent the SSS problem \cite{sssLDA,subspaceLDA,overviewLDA}.

For gFDA, the objective function $f_g$ can be rewritten in the linear combination form of the two symmetrical matrices, as will be proved in Sec.\ref{s:dualforms}. This enables gFDA to avoid the SSS problem and work even with only one sample without any modification. However, in terms of computational cost, it is desirable to use the PCA based dimensionality reduction together, as it can largely reduce the data dimension.
For gFDA, the dimension of the original dimension can be in fact reduced to the number of the orthonormal basis vectors without losing any structural information of the class subspaces, since the orthonormal basis vectors over all the classes are linearly independent, assuming no overlap among class subspaces.  

\subsection{Comparison of FDA and {\bf{g}}FDA}\label{compFDAandgFDA}
Fig.\ref{comp_fda_gfda} shows the comparisons of projections onto discriminant spaces generated by FDA (left) and gFDA (right), where we used sets of face images from the Yale face database. In this database, each subject class contains 45 front face images which were collected under different lighting conditions. 
It is known that all the possible images of a face under various lighting conditions are contained in an illumination cone~\cite{9PL}. 
The illumination cone of a subject can be accurately approximated by a convex cone formed by a set of nine front face images of the subject under nine specific lighting conditions. These nine images are called the 9PL images \cite{9PL} in the Yale face database. Further, the illumination cone is contained in a 9-dimensional illumination subspace, which can be generated by applying PCA to a set of the 9PL images. 
Hence, a 9-dimensional illumination subspace can in principle contain other 36 images under different illumination conditions. 
For more details of the Yale database, see Section~\ref{s:experiments}. 

We used the 9PL images as training data, and used the remaining face images as test data. The dimension of each class subspace was set to 9. In Fig.\ref{comp_fda_gfda}, a row represents the case of 2, 3 or 4 classes. We used FDA with PCA dimensionality reduction \cite{fda2}, since the original FDA cannot be used under this setting due to the SSS problem. In contrast, gFDA avoids the SSS problem by using the linear combination form, which will be described in Sec. \ref{s:dualforms}.


\begin{figure}[bt]
  \begin{center}
    \includegraphics[width=45mm]{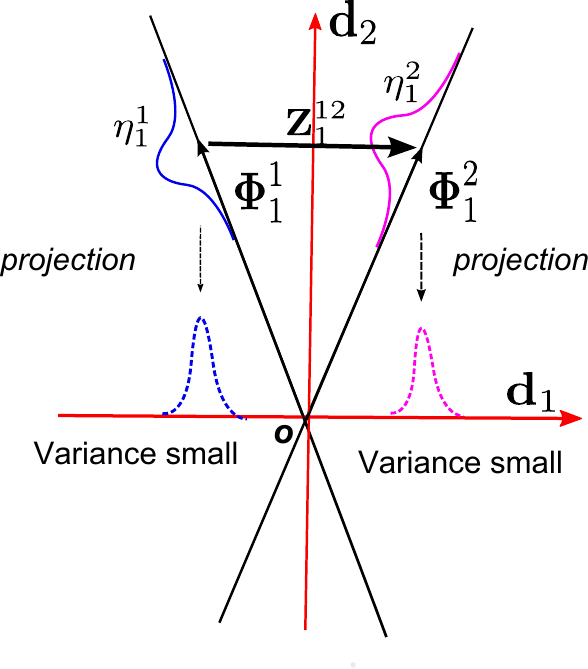}
    \caption{The simplest case: data set of two classes are projected onto a vector $\bf{d}$, respectively. }
    \label{dsprj1}
  \end{center}
\end{figure}

\section{Geometrical mechanism of {\bf{g}}FDA}\label{s:mechanismgFDA}
\subsection{Criterion based on class subspaces}
Our Fisher-like criterion $f_g({\bf{d}})$ is defined by using only the principal component vectors $\{{\bs{\phi}}_i^c\}_{i=1}^{N_c}$. This can be interpreted as that $f_g({\bf{d}})$ is determined based on the geometry of the class subspaces, which are spanned by the principal component vectors $\{{\bs{\phi}}_i^c\}_{i=1}^{N_c}$ of each class $c$.

More specifically, the denominator of $f_g({\bf{d}})$ indicates the sum of the orthogonal projection matrices of all the class subspaces and the numerator the autocorrelation matrix of all the difference vectors among the first orthogonal basis vectors, namely, their mean vectors.  
This indicates that the maximization of $f_g({\bf{d}})$ can be realized, by minimizing the sum of projections of all the class subspaces while maximizing the projections of the differences between the mean vectors at the same time. Reflecting this mechanism, we name the discriminant analysis based on our Fisher-like criterion 
"geometrical FDA (gFDA)".

In the following, we look at  the geometrical characteristics of gFDA through the simplest case that two classes, where their distributions, $\eta_1^1$ and $\eta_1^2$, are on one-dimensional subspaces, ${\cal{P}}_1=\left<{\bs{\phi}}_1^1\right>$, ${\cal{P}}_2=\left<{\bs{\phi}}_1^2\right>$, and ${{\bs{\phi}}_1^1}^{T} {\bs{\phi}}_1^2 >0$. Assume that the two classes have an identical number of data, $n$, and the same variance, and the norms of their mean vectors are 1.0, as shown in Fig.\ref{dsprj1}.

A basis vector ${\bf{d}}_1$ of discriminant space $\cal{F}$ is obtained by solving ${\SB}_3 {\bf{d}}=\lambda {\SW}_4 {\bf{d}}$, where ${\SB}_3 = ({\bs{\phi}}_1^1 - {\bs{\phi}}_1^2) {({\bs{\phi}}_1^1 - {\bs{\phi}}_1^2)}^T$ and ${\SW}_4={\bs{\phi}}_1^1 {{\bs{\phi}}_1^1}^T + {\bs{\phi}}_1^2 {{\bs{\phi}}_1^2}^T$ in this case. 
The ${\bf{d}}_1$ has the same direction as that of ${\bs{\phi}}_1^2 - {\bs{\phi}}_1^1$. Thus, we can see from the geometry shown in Fig.\ref{dsprj1} that the projection onto the discriminant space $\cal{F}$ spanned by ${\bf{d}}_1$ minimizes the sum of projections of the two class subspaces, while maintaining the difference vector, ${\bs{\phi}}_1^2 - {\bs{\phi}}_1^1$. For the data distributions, $\eta_1^1$ and $\eta_1^2$, we can see that the projection reduces the within-class variances, while maintaining the between-class variance. This mechanism can work independently of the type of class distribution, although it may be generally approximated by a normal distribution.
In more general cases with $C (\geq 2)$ classes and multiple dimensions, gFDA still has a similar geometry as in this simple case.

\begin{figure}[bt]
  \begin{center}
\includegraphics[width=90mm]{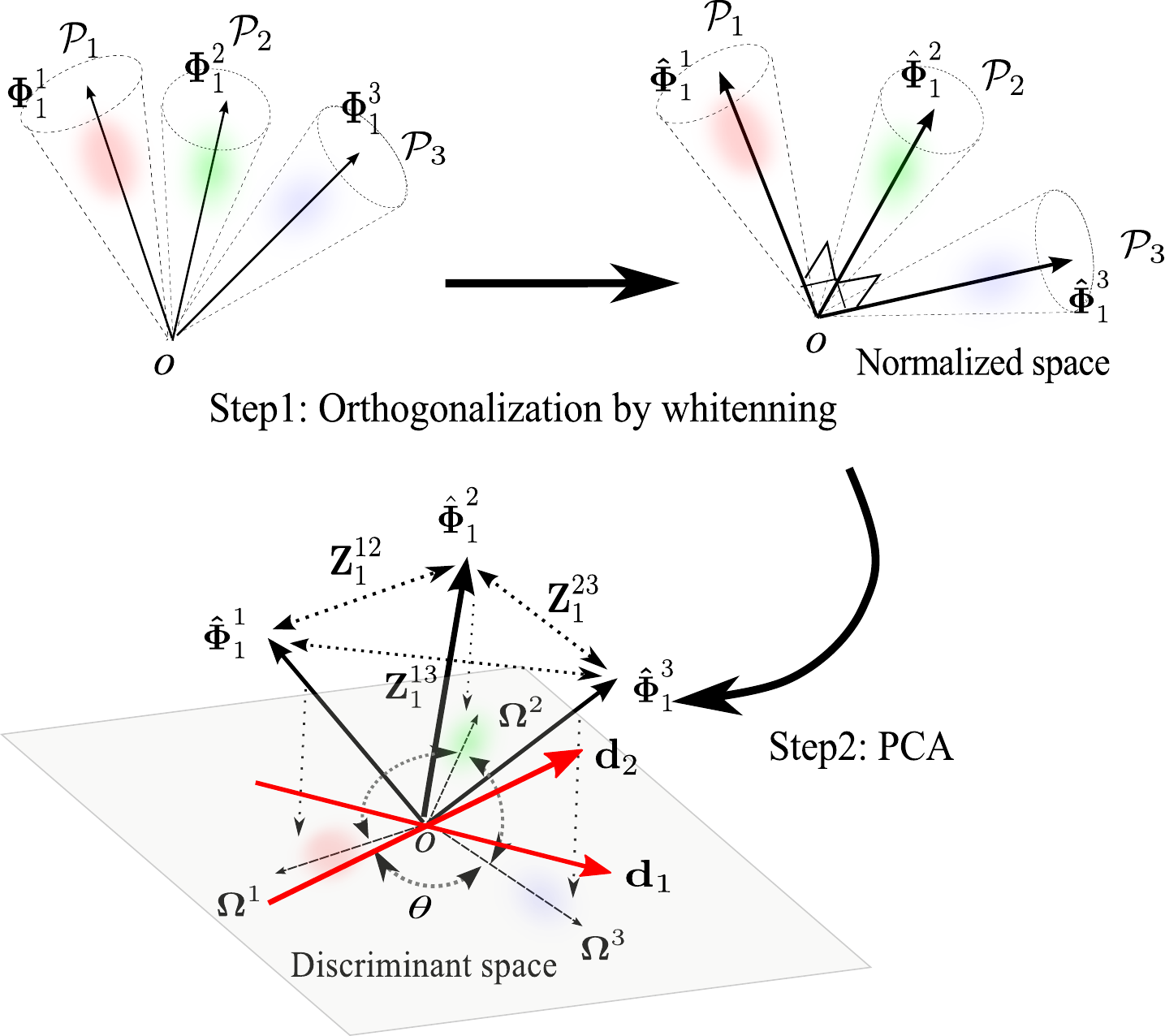}
\caption{Geometrical mechanism of gFDA, which consists of two processes: whitening and PCA. Three mean vectors $\{{\bs{\hat{\phi}}}_1^c\}_{c=1}^3$ are orthogonalized to each other in the normalized space, and they are then projected onto the discriminant space.}
\label{gfda_flow3}
  \end{center}
\end{figure}

\subsection{Two-steps process}
It is well known that the whole process of FDA consists of two steps: whitening and PCA. The process of gFDA in the form of $\frac{{\bf{d}}^T{\SB}_3{\bf{d}}}{{\bf{d}}^T{\SW}_4{\bf{d}}}$ can be also divided into these two steps as shown in Fig.\ref{gfda_flow3}.

We consider the case that $C$ $N$-dimensional class subspaces in ${\bf{R}}^{L}\, (L \gg N)$ are given, assuming that there is no overlap between class subspaces. 
For the simplicity of discussion, to make the matrix ${{\SW}_4}$ full rank, we assume that the dimensionality of the vector space can be reduced from $L$ to $\hat{L}=C N$ by applying PCA-based dimension reduction. Thus, in the following, we consider $C$ $N$-dimensional class subspaces in ${\bf{R}}^{{\hat{L}}}$. 
The details of each step are as follows:%
\begin{itemize}
\item[1)] In the first step, whitening $\bf{A}$ such that ${\bf{A}}^T{\SW}_4{\bf{A}}={\bf{E}}$ is applied to $C N$ orthonormal basis vectors $\{{\bs{\phi}}_i^c\}$ of $C$ $N$-dimensional class subspaces. As a result, the orthonormal basis vectors of all the classes are orthogonalized to each other. A subspace spanned by these orthogonalized basis vectors in the first step is called normalized space in contrast with the original feature vector space. Let the orthogonalized basis vectors be $\{{\bs{\hat{\phi}}}_i^c\}$ in the normalized space.
\item[2)] In the second step, PCA is applied to a set of the difference vectors $\{{\bf{z}}_1^{ij}\}$$(i=1,\dots,C-1,j=i+1,\dots,C)$ between the first principal component vectors, $\{{\bs{\hat{\phi}}}_1^c\}_{c=1}^C$, where ${\bf{z}}_1^{ij}$=${\bs{\hat{\phi}}}_1^i - {\bs{\hat{\phi}}}_1^j$, where ${{\bs{\hat{\phi}}}_1^i}^T {\bs{\hat{\phi}}}_1^j > 0$. We obtain $C-1$ principal component vectors $\{{\bf{d}}_i\}_{i=1}^{C-1}$ from ${\bf{\Sigma}}_A=\sum_{i=1}^{C}\sum_{j=i+1}^C {\bf{z}}_1^{ij}{{\bf{z}}_1^{ij}}^T$, since the rank of ${\bf{\Sigma}}_A$ is $C-1$. Note that ${\bf{\Sigma}}_A$ can be also represented by ${\bf{A}}^T {\SB}_{3}{\bf{A}}$.
\end{itemize}
The obtained principal component vectors $\{{\bf{d}}_i\}_{i=1}^{C-1}$ span the discriminant space $\cal{F}$.

\subsection{Dual forms of objective function} \label{s:dualforms}
The objective function $f_g$ of our simplified Fisher criterion is represented as a generalized eigenvalue problem for the matrix product ${{\SW}_4}^{-1} {\SB}_3$. In the following, we prove that the objective function can also be represented as a simpler regular eigenvalue problem for the linear combination of ${\SB}_3$ and ${\SW}_4$ under the same setting as in the previous section. We consider a set of $C$ $N$-dimensional class subspaces in ${\bf{R}}^{{\hat{L}}}$. 


The flow of our proof is summarized as follows:
\begin{itemize}
\item[C1.] $C-1$ eigenvalues of matrix ${{\SW}_4}^{-1} {\SB}_3$ $\in {\bf{R}}^{{\hat{L}}{\times}{\hat{L}}}$ are all equal to $C$ without depending on the dimensionality of each class subspace. 
\item[C2.] The characteristic C1 above leads to the following equivalent relationship:\\
${{\SW}_4}^{-1}{\SB}_3{\bf{d}}= C {\bf{d}}~
\Leftrightarrow~
({\SW}_4-\frac{1}{C}{\SB}_3){\bf{d}}=0 {\bf{d}}$, \\
where we note that in the former equation we need to take 
the eigenvectors 
corresponding to $C-1$ largest eigenvalues, while in the latter we need to take the eigenvectors corresponding to $C-1$ smallest eigenvalues (zero).
\end{itemize}
The two sets of the eigenvectors obtained from the two eigenvalue problems in C2 are different. In fact, the former eigenvectors are not orthogonal, since the matrix ${{\SW}_4}^{-1}{\SB}_3$ is not symmetrical. In contrast, the latter eigenvectors 
are orthogonal to each other, since matrix ${\SW}_4-\frac{1}{C}{\SB}_3$ is symmetrical.
However, the two subspaces spanned by the respective sets of the eigenvectors 
coincide completely. 
Therefore, we will confirm that gFDA has dual forms of objective function. 

\vspace{2mm}
\noindent{\bf{Proof of C1}}.
The characteristic C1 can be proved as follows: 
${{\SW}_4}^{-1}{\SB}_3$ has the same eigenvalues as ${\bf{\Sigma}}_A = {\bf{A}}^T{\SB}_3 {\bf{A}}$, where $\bf{A}$ is the whitening such that ${\bf{A}}^T {\SW}_4 {\bf{A}}={\bf{E}}$, as described in the previous section. ${\bf{\Sigma}}_A$ is represented with the difference vectors between the $C$ first orthonormal basis vectors, $\{{\bs{\hat{\phi}}}_1^c\}$, which are orthogonalized by whitening $\bf{A}$:
\begin{eqnarray}
{\bf{\Sigma}}_A = \sum_{i,j=1,i<j}^C ({{\bs{\hat{\phi}}}_1^i}-{{\bs{\hat{\phi}}}_1^j}) {({{\bs{\hat{\phi}}}_1^i}-{{\bs{\hat{\phi}}}_1^j})}^T.
\end{eqnarray}
Let ${\hat{\bf{\Sigma}}}_A$ $\in {\bf{R}}^{\hat{L}{\times}\hat{L}}$ be the autocorrelation matrix of the difference vectors among the standard basis $\{{\bf{e}}_1,\dots,{\bf{e}}_C\}$ of ${\bf{R}}^{\hat{f}}$:
\begin{equation}
{\hat{\bf{\Sigma}}}_A=\sum_{i,j=1,i<j}^C ({\bf{e}}_i-{\bf{e}}_j) {({\bf{e}}_i-{\bf{e}}_j)}^T.
\end{equation}
Since both the basis of ${\{{\bs{\hat{\phi}}}_1^c\}}_{c=1}^C$ and the standard basis of $\{{\bf{e}}_1,\dots,{\bf{e}}_C\}$ are orthonormal basis of ${\bf{R}}^{\hat{L}}$, the two autocorrelation matrices ${\bf{\Sigma}}_A$ and ${\hat{\bf{\Sigma}}}_A$ have the same eigenvalues, though the corresponding eigenvectors are different. 

${\hat{\bf{\Sigma}}}_A$ can be written as
\begin{equation}
{\hat{\bf{\Sigma}}}_A=
\left(\begin{array}{cccc}
C-1 & -1 & \cdots & -1 \\
-1 & C-1  & \cdots & -1 \\
\vdots & \vdots & \ddots & \vdots \\
-1  & -1 & \cdots & C-1
\end{array}
\right)
\end{equation}
\begin{equation}
=
C \left(\begin{array}{cccc}
1 & 0 & \cdots & 0 \\
0 & 1  & \cdots & 0 \\
\vdots & \vdots & \ddots & \vdots \\
0  & 0 & \cdots & 1
\end{array}
\right)
-
\left(\begin{array}{cccc}
1 & 1 & \cdots & 1 \\
1 & 1  & \cdots & 1 \\
\vdots & \vdots & \ddots & \vdots \\
1  & 1 & \cdots & 1
\end{array}
\right).
\end{equation}
In the above equation, the first matrix has $C$ eigenvalues of $C$ and the second one has one $C$ and $C-1$ zeros as the eigenvalues. Hence, matrix ${\hat{\bf{\Sigma}}}_A$ has $C-1$ eigenvalues of $C$ as non-zero eigenvalue. Therefore, we can confirm that ${\bf{\Sigma}}_A$ has $C-1$ eigenvalues of $C$ as well.

\vspace{2mm}
\noindent{\bf{Proof of C2}}.
Next, we shall prove characteristics C2. By substituting $\lambda=C$ into Eq.(\ref{fg2}), we obtain
\begin{eqnarray}
{\SB}_3{\bf{d}}=C{{\SW}_4}{\bf{d}} \label{eq1}.
\end{eqnarray}
Further, we can rewrite the equation as
\begin{eqnarray}
({{\SW}_4}-\frac{1}{C}{\SB}_3){\bf{d}}=0= 0{\bf{d}}, \label{eq2}
\end{eqnarray}
where, by considering that $\bf{d}$ is not a zero vector, ${{\SW}_4}-\frac{1}{C}{\SB}_3$ has $C-1$ zero eigenvalues. 

This characteristic means that the eigenspace (solution space) of ${{\SW}_4}-\frac{1}{C}{\SB}_3$ corresponding to zero eigenvalue is equivalent to that of ${{\SW}_4}^{-1}{\SB}_3$ corresponding to the eigenvalue of $C$.
In other words, the discriminant space $\cal F$ spanned by the $C-1$ eigenvectors of ${{\SW}_4}-\frac{1}{C}{\SB}_3$ corresponding to zero eigenvalues coincides with that spanned by the $C-1$ eigenvectors of $ {{\SW}_4}^{-1}{\SB}_3$. In the following, we use ${\hat{\bf{G}}}$ to indicate ${\SW}_{4} -\frac{1}{C}{\SB}_{3}$.

Thus far, we considered the vector space with the reduced dimension 
of $\hat{L}$. However, we should note that the above discussion can hold so that the linear combination form is also valid in the original vector space with the dimension 
of $L$. 

Here, we reiterate that the linear combination form can mitigate the SSS problem, since it can be stably calculated independently of the number of data and the dimension of vector space, unlike the matrix product form.

\subsection{Generation of discriminant space}
As stated in the previous section, all of $C-1$ discriminant vectors ${\{{\bf{d}}_i\}}_{i=1}^{C-1}$ have the same discriminant ability, $C$. This characteristic suggests that each individual vector of 
$C-1$ 
${\bf{d}}_i$ does not have a meaning, rather a subspace spanned by them should be considered to be essential. Hence, we define a subspace spanned by $C-1$ discriminant vectors as discriminant space $\cal{F}$, where the discriminant vectors are orthogonalized to each other by using the Gram-Schmidt orthonormalization.

\begin{figure}
\begin{center}
\includegraphics[width=85mm]{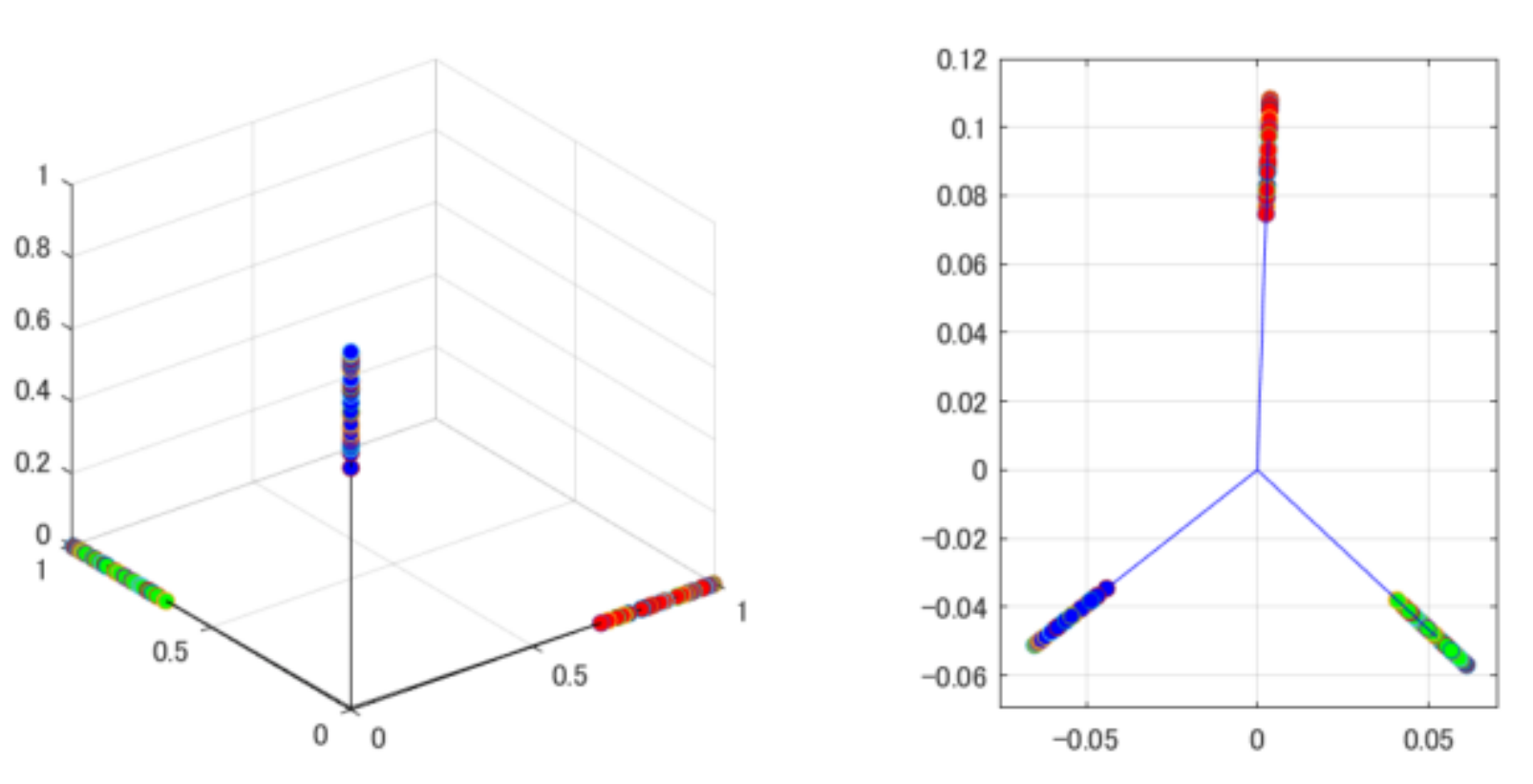}\\
(a) Normalized space \hspace{10mm} (b) Discriminant space\\
\includegraphics[width=85mm]{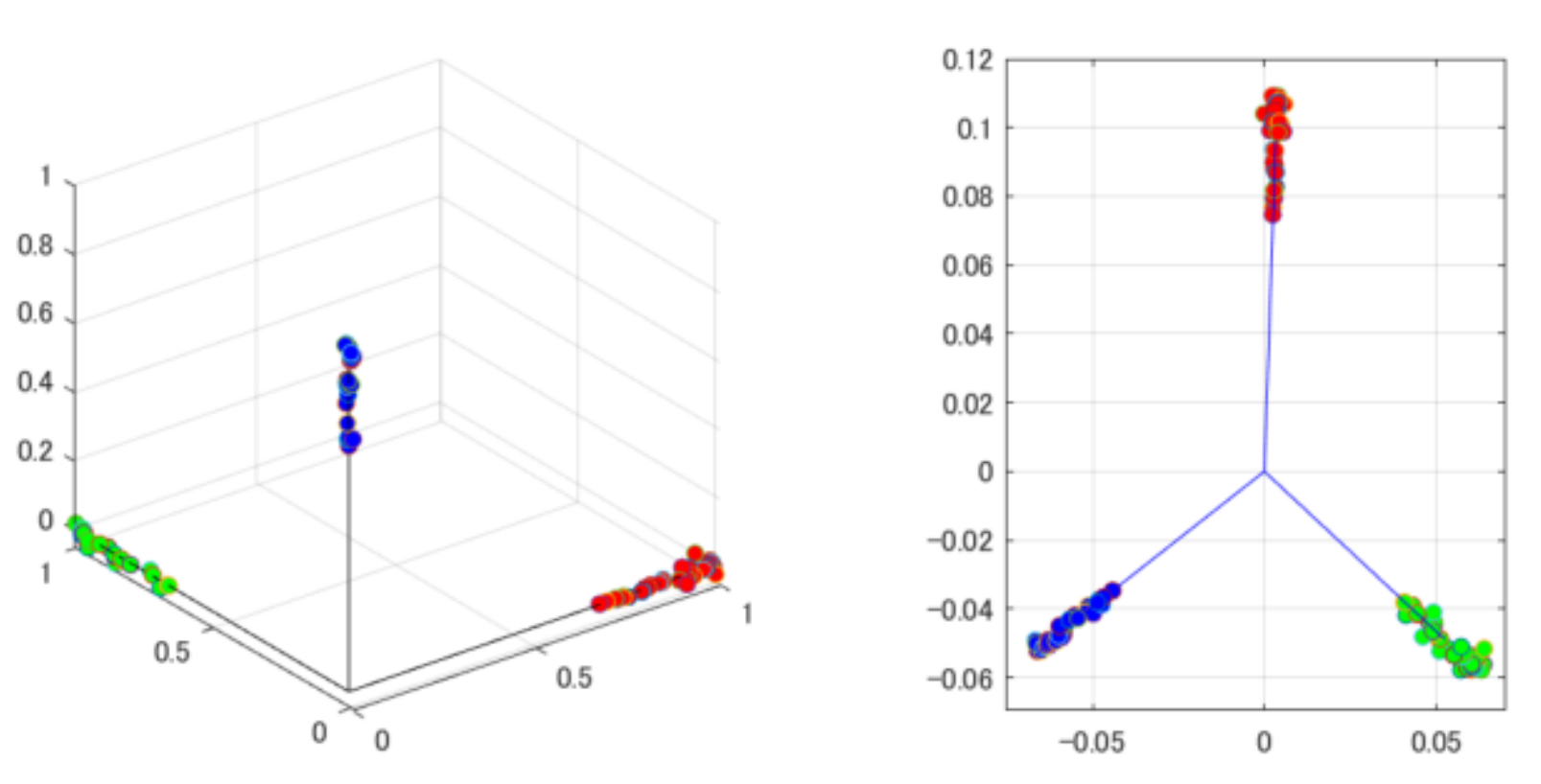}\\
(c) Normalized space \hspace{10mm} (d) Discriminant space
\caption{Effectiveness of normalization: (a) and (b) show the projections of data of 3 classes in the normalized spaces and discriminant space, respectively, where all data of each class are completely contained within its class subspace. (c) and (d) show the projections of data of the 3classes, where some component of data are not covered by the class subspaces.} 
\label{normalization}
\end{center}
\end{figure}

\subsection{Effectiveness of normalization}\label{effectnessNorm}
To get the maximum performance out of gFDA in a classification task, we essentially need to incorporate the normalization of orthogonal projection $\tau({\bf{x}}) \in {\bf{R}}^{C-1}$ of data ${\bf{x}} \in {\bf{R}}^L$ on the discriminant space $\cal{F}$ into the mechanism of gFDA, where  the normalization is defined as ${\tau({\bf{x}})}/||{\tau({\bf{x}})} ||$. 

In the following, we discuss the reason for this from the viewpoint of geometrical structure of gFDA.
In a normalized space, only the first basis vectors $\{{\bs{\hat{\phi}}_1^c}\}$ 
are selected from the orthogonalized basis vectors ${\{\bs{\hat{\phi}}}_i^c\}$ of all the class subspaces, and the remaining basis vectors ${\{\bs{\hat{\phi}}_i^c\}}_{i=2}^{d_c}$ are discarded.
This operation results in that the data of class $c$ are projected onto only the $\{{\bs{\hat{\phi}}_1^c}\}$ in the normalized space as shown in Fig.\ref{normalization}a, when all the data of class $c$ are completely contained within the $c$-th class subspace spanned by ${\{\bs{{\phi}}}_i^c\}_{i=1}^{d_c}$. Such a situation corresponds to that the illumination subspace of an object contains any images of the object under various illumination conditions as described in Sec.\ref{compFDAandgFDA}.
However, as it is in general difficult to generate such an illumination subspace in practical applications, the projected data points of the $c$-the class in the normalized space can have nonzero components on the basis vectors $\{{\bs{\hat{\phi}}_1^{c'}}\} (c' \neq c)$ of other classes. 
As a result, they are projected at a remove from $\{{\bs{\hat{\phi}}_1^c}\}$ as shown in Fig.\ref{normalization}c. This geometrical relationship remains in the discriminant space $\cal{F}$ as shown in Figs.\ref{normalization}b and d. 

We should note that in the above process, the variation of the projections in the direction of $\{{\bs{\hat{\phi}}_1^c}\}$ can necessarily remain even if we could generate the illumination subspace of class $c$. Namely, we cannot in principle remove them.
A valid way for ignoring this extra variation is to measure the similarity between the projections of an orthogonalized input $\hat{\bf{x}}$ and  $\{{\bs{\hat{\phi}}_1^c}\}$ onto the discriminant space $\cal{F}$ by the angle $\theta$ between them.
Here, we need to use $cos \theta$ instead of $cos^2 \theta$ as a similarity in order to deal with the cases that the angle between $\hat{\bf{x}}$ and ${\bs{\hat{\phi}}_1^c}$ is over 90 degrees, i.e., ${\hat{\bf{x}}}^T {\bs{\hat{\phi}}_1^c} < 0$.
This angle based classification corresponds to the Euclidean distance based classification with the normalization according to the cosine theorem. 

\section{Connection of {\bf{g}}FDA and GDS projection} \label{s:connection}
In this section, we show a close connection between gFDA and GDS projection. To this end, we prove that gFDA is equivalent to GDS projection with a small correction item.

\begin{figure}[bt]
  \begin{center}
\includegraphics[width=84mm]{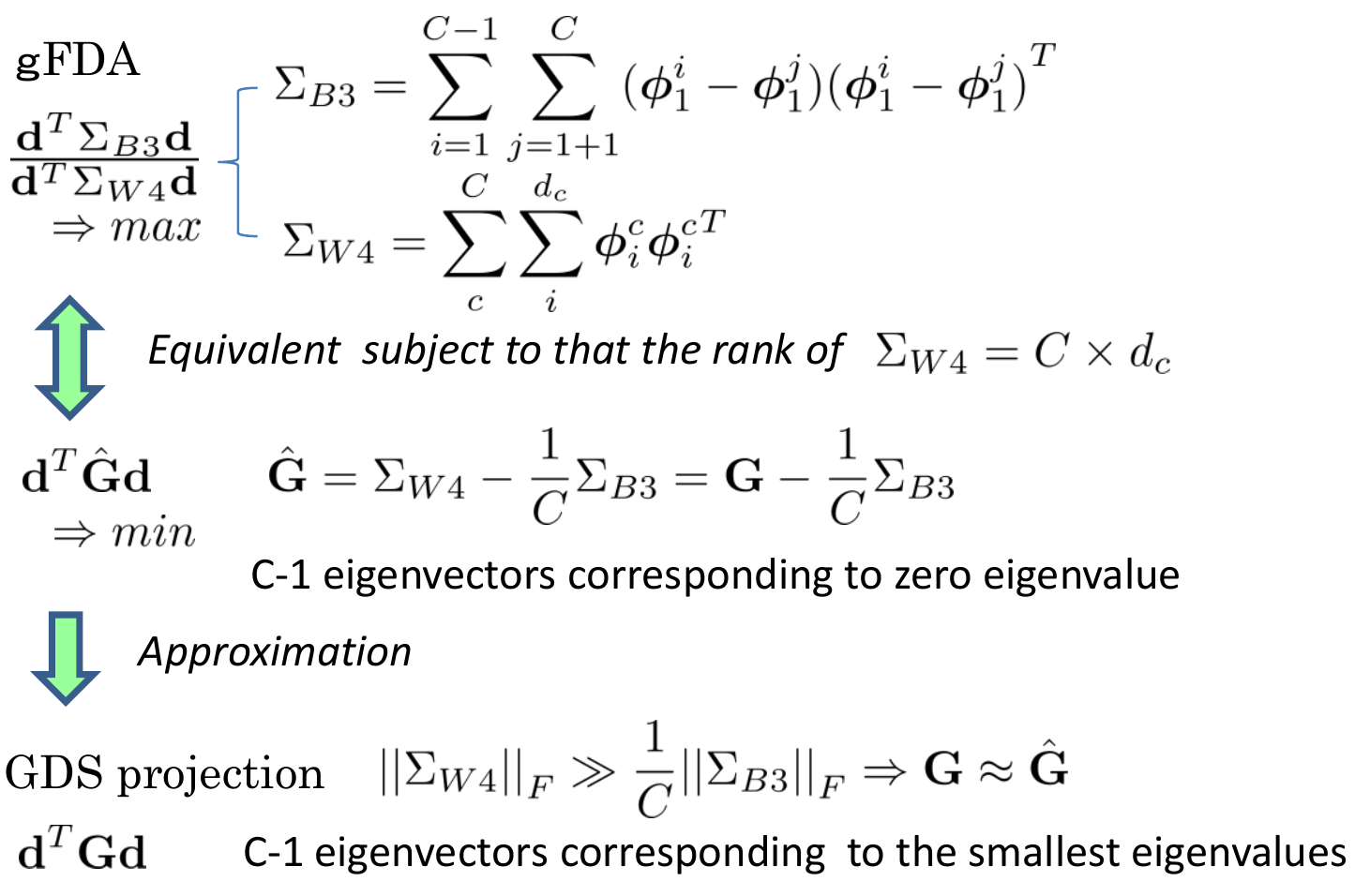}
\caption{Connection between gFDA and GDS projection.}
\label{fromgFDAtoGDS}
  \end{center}
\end{figure}

\begin{figure}[bt]
  \begin{center}
\subfloat[gFDA]{\includegraphics[width=44mm]{./gfda_yale_2classes.pdf}}
\subfloat[GDS]{\includegraphics[width=39mm]{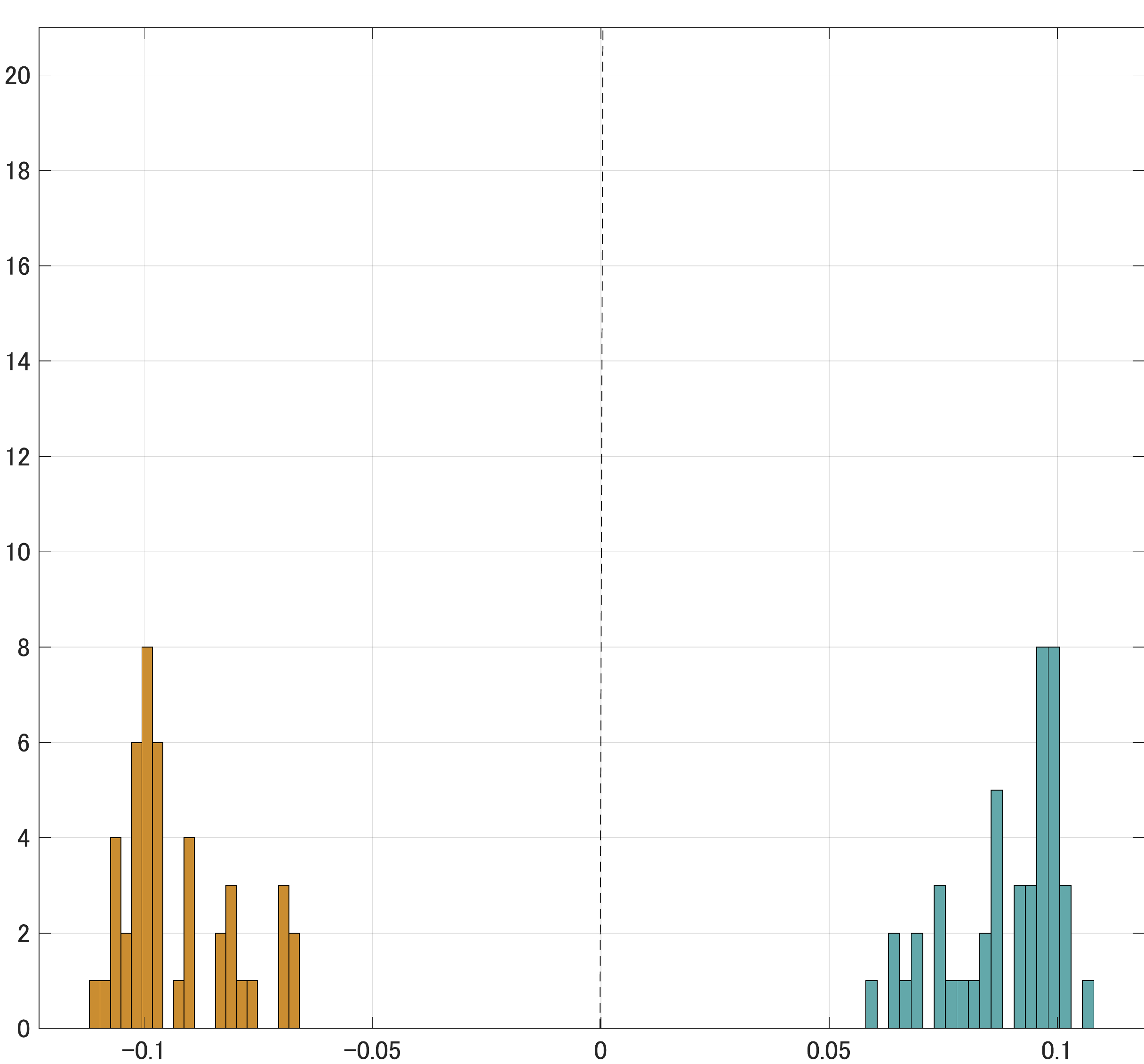}}\\
\subfloat[gFDA]{\includegraphics[width=39mm]{./gfda_yale_3classes0610.pdf}}~~
\subfloat[GDS]{\includegraphics[width=39mm]{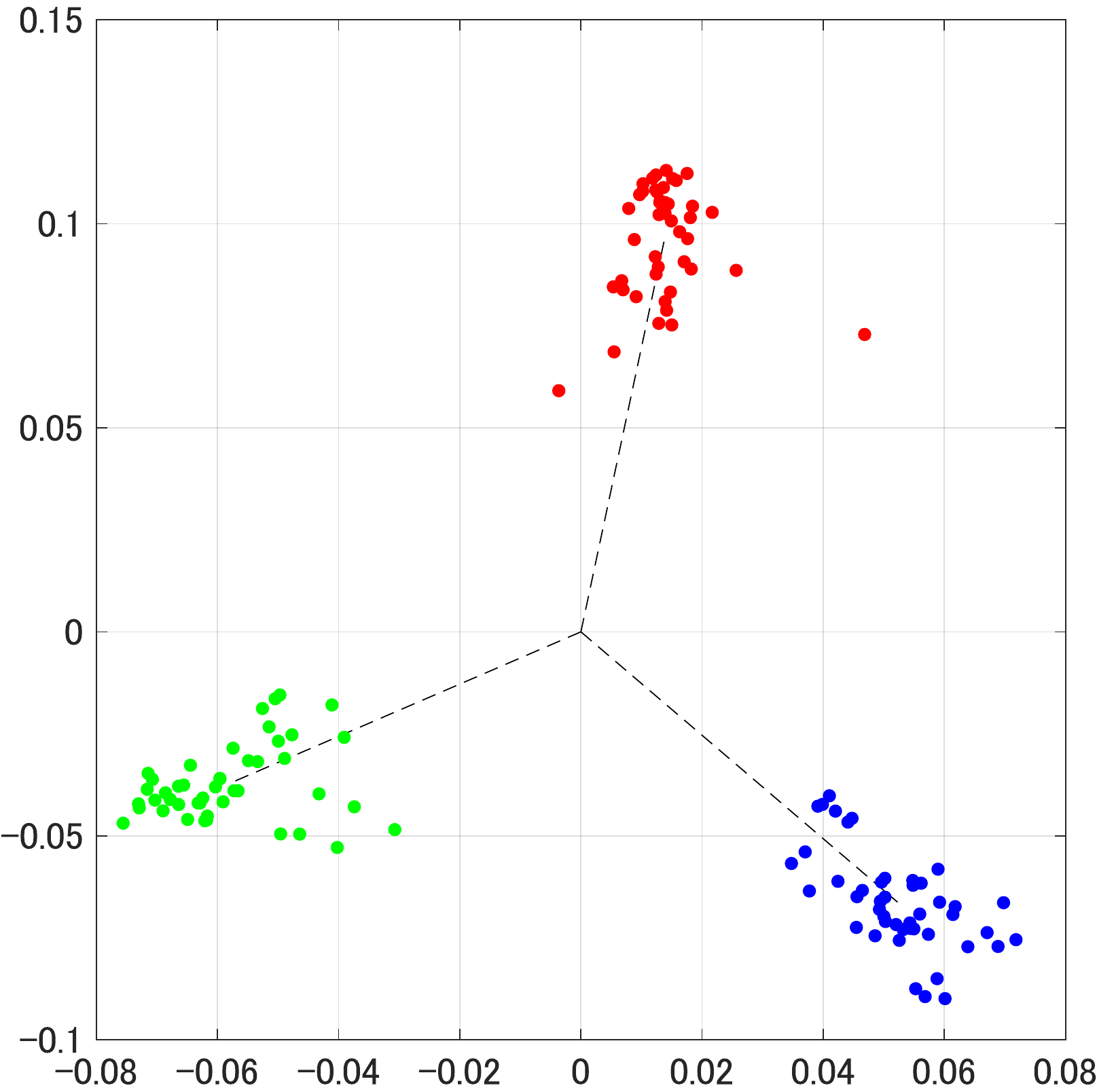}}\\
\subfloat[gFDA]{\includegraphics[width=45mm]{./gfda_yale_4classes0611.pdf}}
\subfloat[GDS]{\includegraphics[width=42mm]{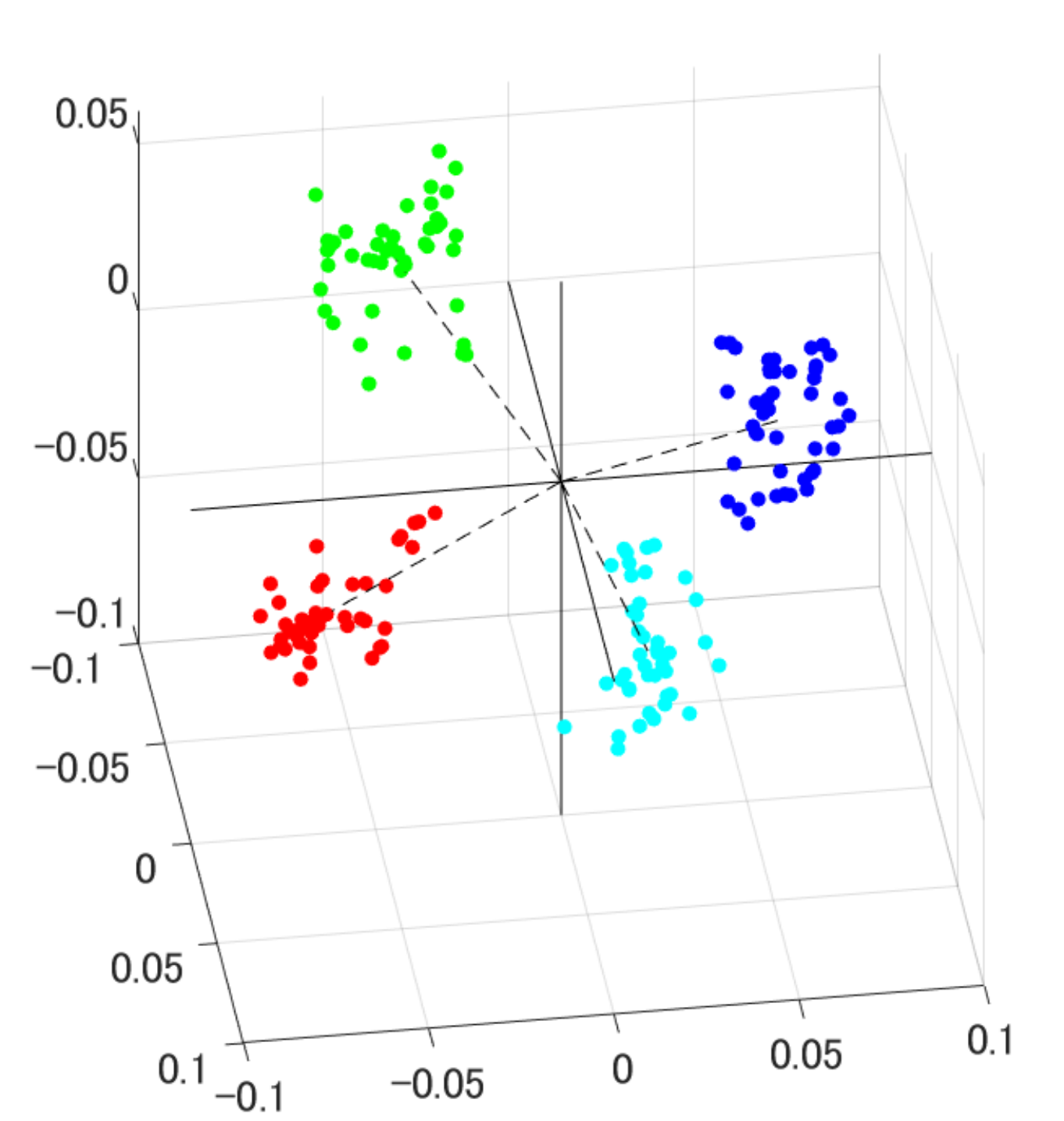}}\\
\caption{Visualization of the projects by gFDA and GDS projection in the cases of two, three and four classes from the Yale face database. }
\label{comp_gfda_gds}
  \end{center}
\end{figure}

\subsection{GDS projection with a small correction term}
According to the new form of ${\SW}_{4} -\frac{1}{C}{\SB}_{3}$ for gFDA in the previous section, we notice that gFDA is closely related to GDS projection \cite{gds} that uses $C-1$ smallest eigenvector of only ${\SW}_4$, 
because given
${||{\SW}_4||}_{F} \gg \frac{1}{C}{||{\SB}_3||}_{F}$. From the standpoint of GDS projection, $\frac{1}{C}{||{\SB}_3||}$ can be regarded as a small correction on itself. Thus, we can regard gFDA as GDS projection with a small correction term of $\frac{1}{C}{\SB}_3$.
Fig.\ref{fromgFDAtoGDS} summarizes the whole flow of the simplification from gFDA to GDS projection that has been discussed so far. The close connection suggests that GDS projection has a discriminant ability and the robustness against the SSS problem as well as gFDA. Fig.\ref{comp_gfda_gds} shows the comparison between gFDA and GDS projection on the examples that were used for the comparison of FDA and gFDA in Fig.\ref{comp_fda_gfda}. We can see high similarity between the results of these two methods.

\subsection{Geometry gap between {\bf{g}}FDA and GDS}
We now discuss the relationship between gFDA and GDS projection in more detail. For this purpose, we introduce a pair of vectors, ${\bf{z}}_{i}^{jk}$ and ${\bf{z'}}_{i}^{jk}$, between the $i$-th orthonormal basis vectors $\ph{i}{j}$ and $\ph{i}{k}$ of classes $j$ and $k$, where ${\bf{z}}_i^{jk} = \ph{i}{j}-\ph{i}{k}$ and ${\bf{z'}}_i^{jk} = \ph{i}{j}+\ph{i}{k}$. Note that ${\bf{z}}_{i}^{jk}{{\bf{z}}_{i}^{jk}}^{T}+{\bf{z'}}_{i}^{jk}{{\bf{z'}}_{i}^{jk}}^{T}={2}(\ph{i}{j}{\ph{i}{j}}^{T}+\ph{i}{k}{\ph{i}{k}}^{T})$.


With $\{\bf{z}\}$ and $\{\bf{z'}\}$,  we rewrite matrix ${\bf{G}}(={\SW}_4)=\sum_{j=1}^{C}\sum_{i=1}^N {\bs{\phi}}_i^j {{\bs{\phi}}_i^j}^{T}$ in Eq.(\ref{defofG}) for GDS projection as follows:
\begin{eqnarray}
{\bf{G}} &=& \frac{1}{2(C-1)}\sum_{j,k, j<k}^C \sum_{i=1}^N ({\bf{z}}_i^{jk}{{\bf{z}}_i^{jk}}^{T}+{\bf{z'}}_i^{jk}{{\bf{z'}}_i^{jk}}^{T}) \label{gds_form1}\\
&=& \frac{1}{2(C-1)}{\SB}_{3} + {\SW}_{5}, \label{gds_form2}
\end{eqnarray}
where 
\begin{eqnarray}
{\SW}_5 = \frac{1}{2(C-1)}\sum_{j,k, j<k}^C {\bf{z'}}_1^{jk}{{\bf{z'}}_1^{jk}}^{T} + \sum_{j=1}^C \sum_{i=2}^{N}  {\bs{\phi}}_i^j  {{\bs{\phi}}_i^j}^T. 
\nonumber
\end{eqnarray}
Eq.(\ref{gds_form1}) indicates that finding the smallest eigenvalues of ${\bf{G}}$ can be regarded as the minimization problem on the sum of the projections of all the vectors $\{{\bf{z}}\}$ and $\{{\bf{z'}}\}$. 
In the same way, we rewrite ${\hat{\bf{G}}}$ for gFDA as follows:
\begin{eqnarray}
{\hat{\bf{G}}} = (\frac{1}{2(C-1)}-\frac{1}{C})
{\SB}_{3}
 + {\SW}_{5}. \label{gfda_form}
\end{eqnarray}

We notice that only the weights on ${\SB}_3$, that is, on the difference vectors $\{{\bf{z}}_1^{jk}\}$), are different between Eq.(\ref{gds_form2}) and Eq.(\ref{gfda_form}), which are $\frac{1}{2(C-1)}>0$ and $\frac{1}{2(C-1)}-\frac{1}{C} \leq 0$, respectively. This difference in the weights produces a geometrical gap between gFDA and GDS projection.
We measure the gap by using an index $\sigma$, which is defined as follows:
\begin{eqnarray}
\sigma = \frac{\frac{1}{2(C-1)} -(\frac{1}{2(C-1)}-\frac{1}{C})}{\frac{1}{2(C-1)}}
=2(1-\frac{1}{C}).
\end{eqnarray}
The value of $\sigma$ becomes larger from 1.0 in the case of  C=2 toward 2.0 as 
the class number 
$C$ increases. Thus, we can see that the gap increases as $C$ gets larger.

\begin{figure}[bt]
  \begin{center}
\subfloat[Eigenvalue (3classes)]{\includegraphics[width=43mm]{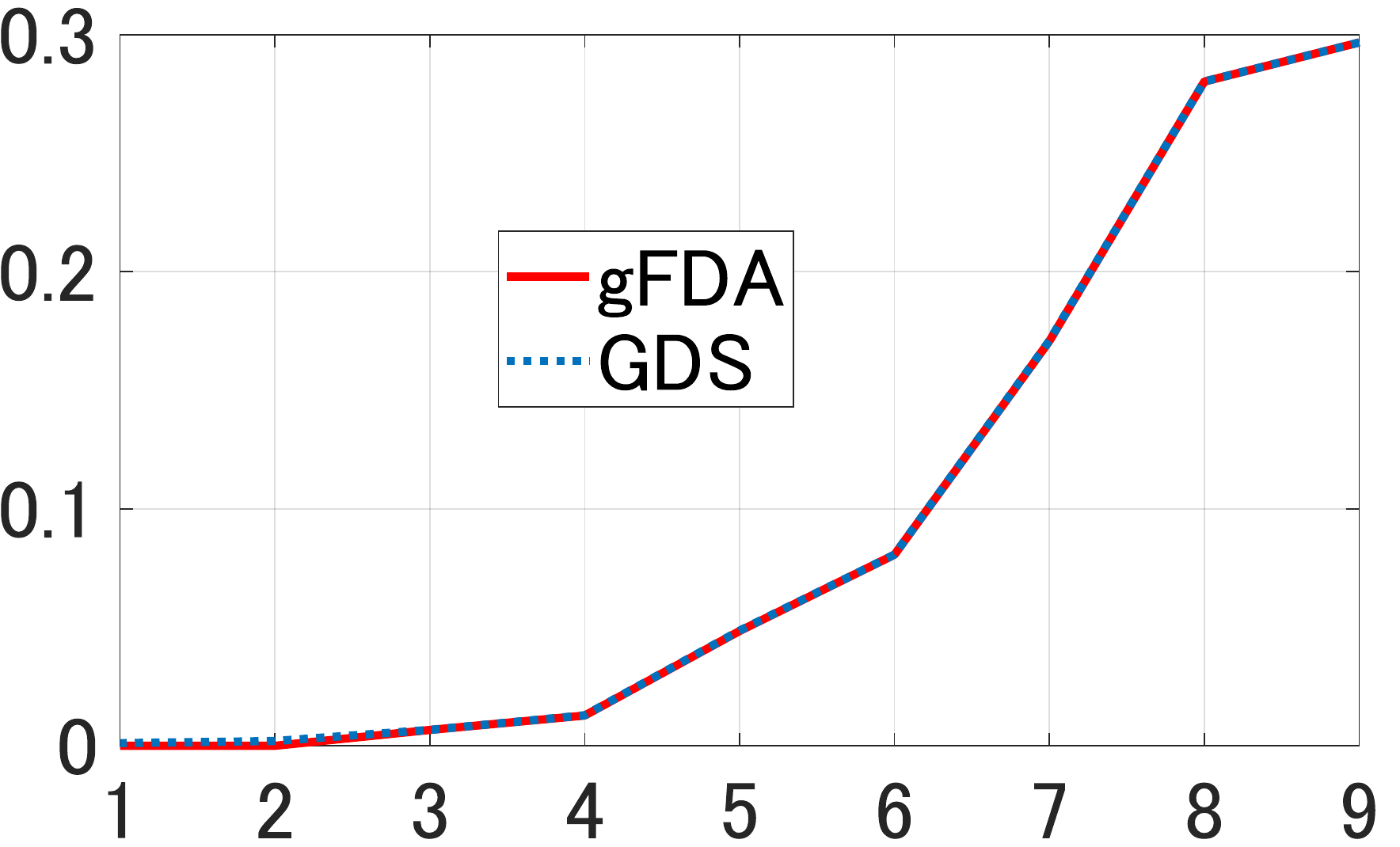}}
\subfloat[F-criterion (3classes)]{\includegraphics[width=43mm]{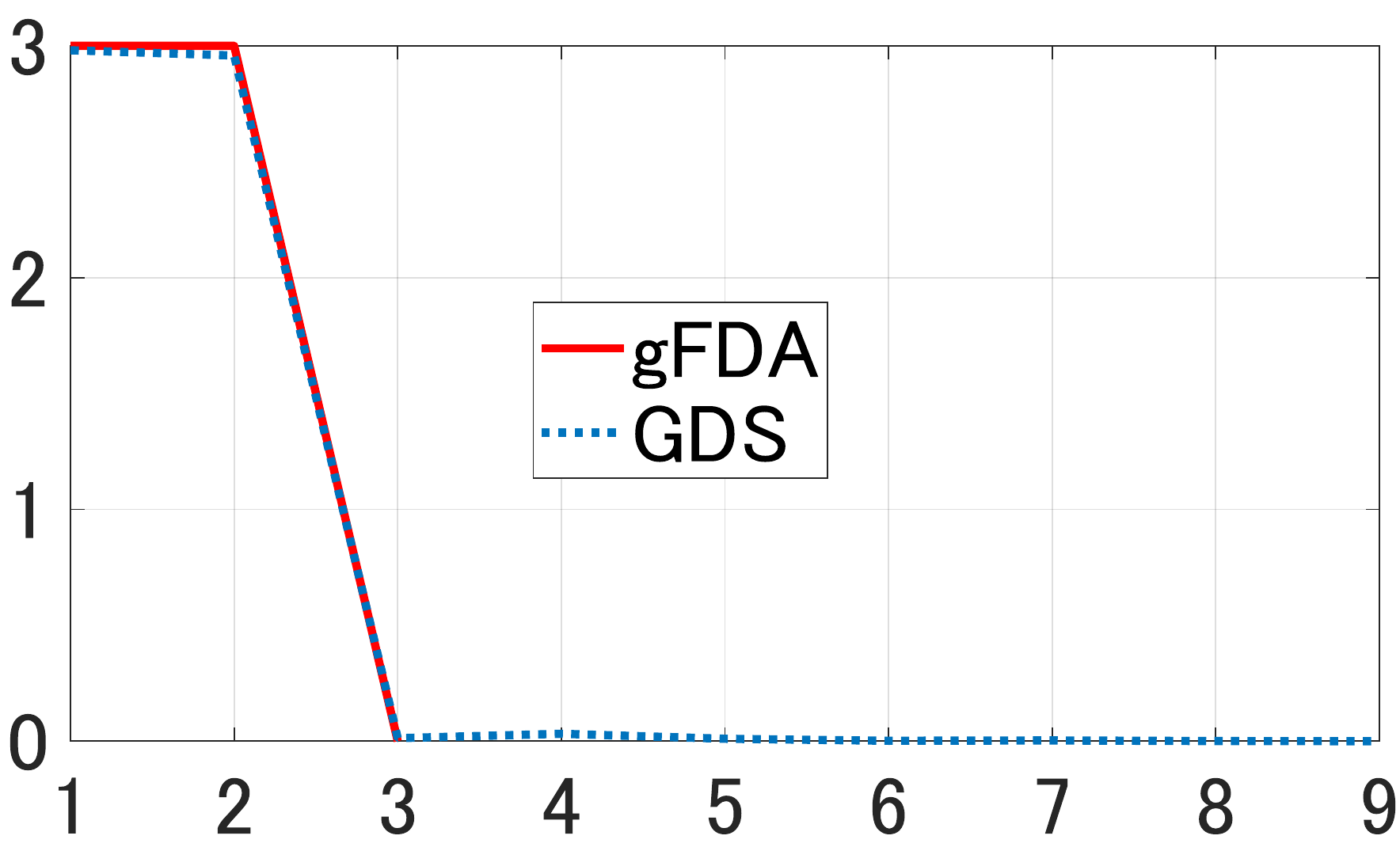}}\\
\subfloat[Eigenvalue (5classes)]{\includegraphics[width=43mm]{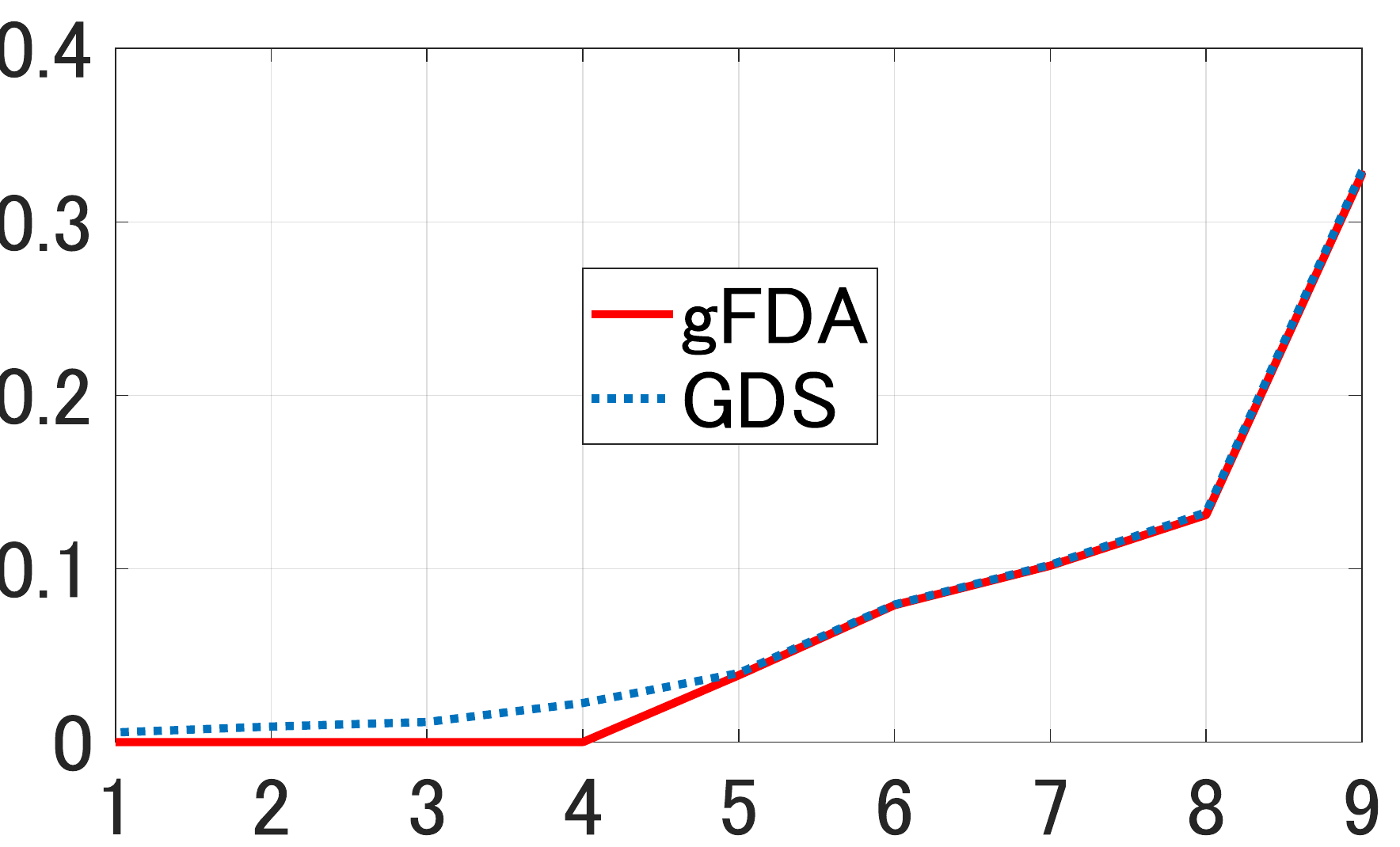}}
\subfloat[F-criterion (5classes)]{\includegraphics[width=43mm]{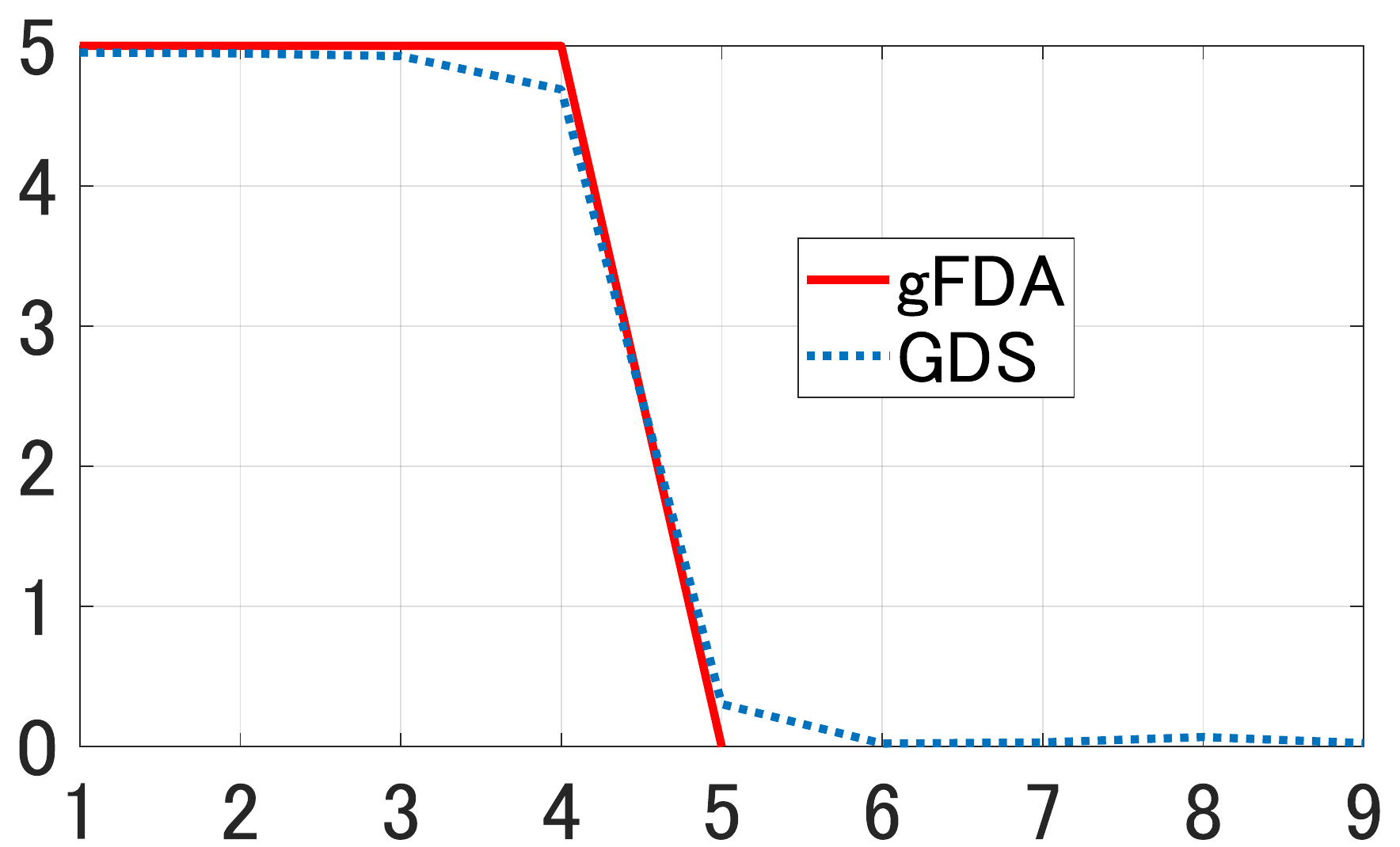}}\\
\subfloat[Eigenvalue (20classes)]{\includegraphics[width=43mm]{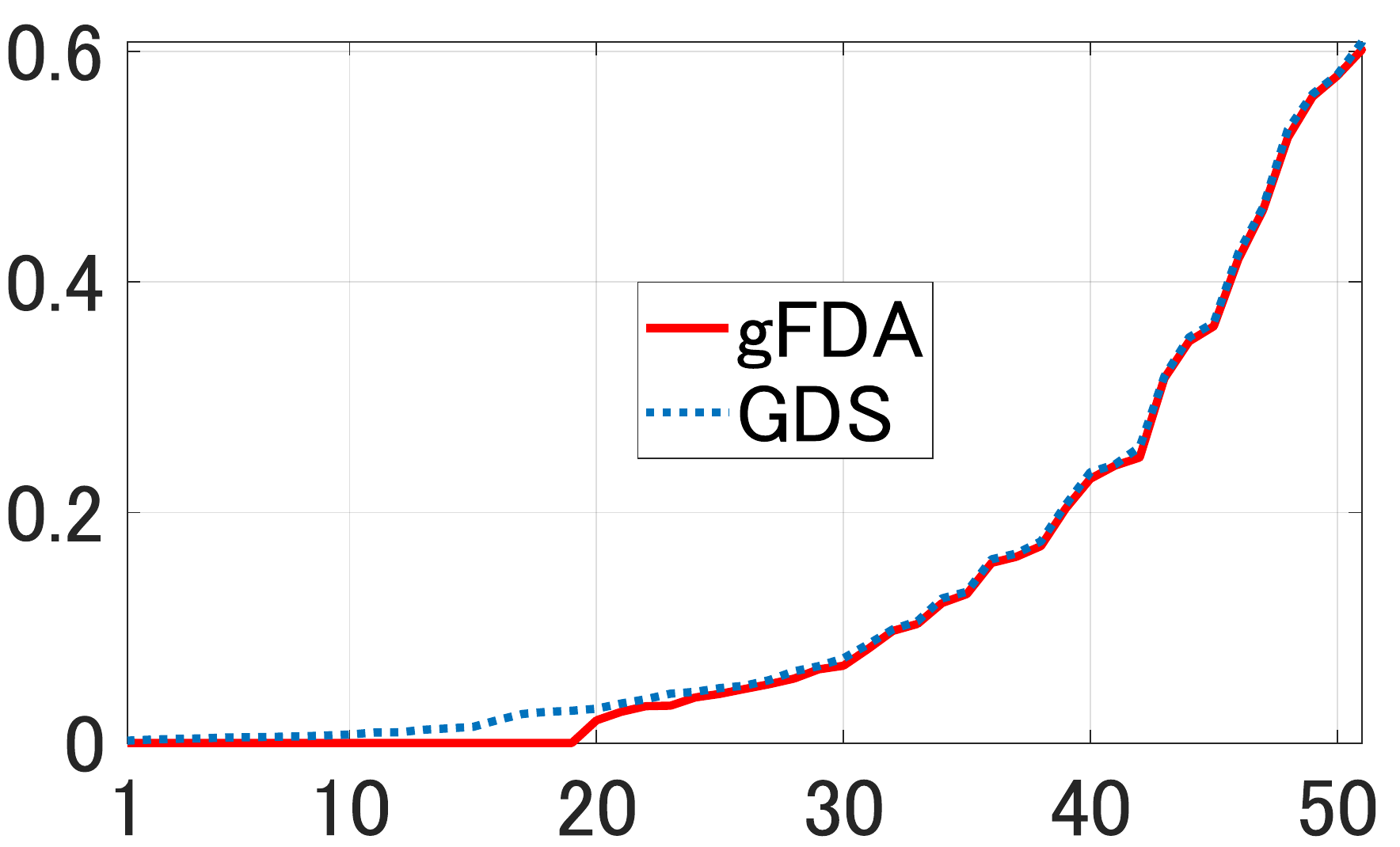}}
\subfloat[F-criterion (20classes)]{\includegraphics[width=43mm]{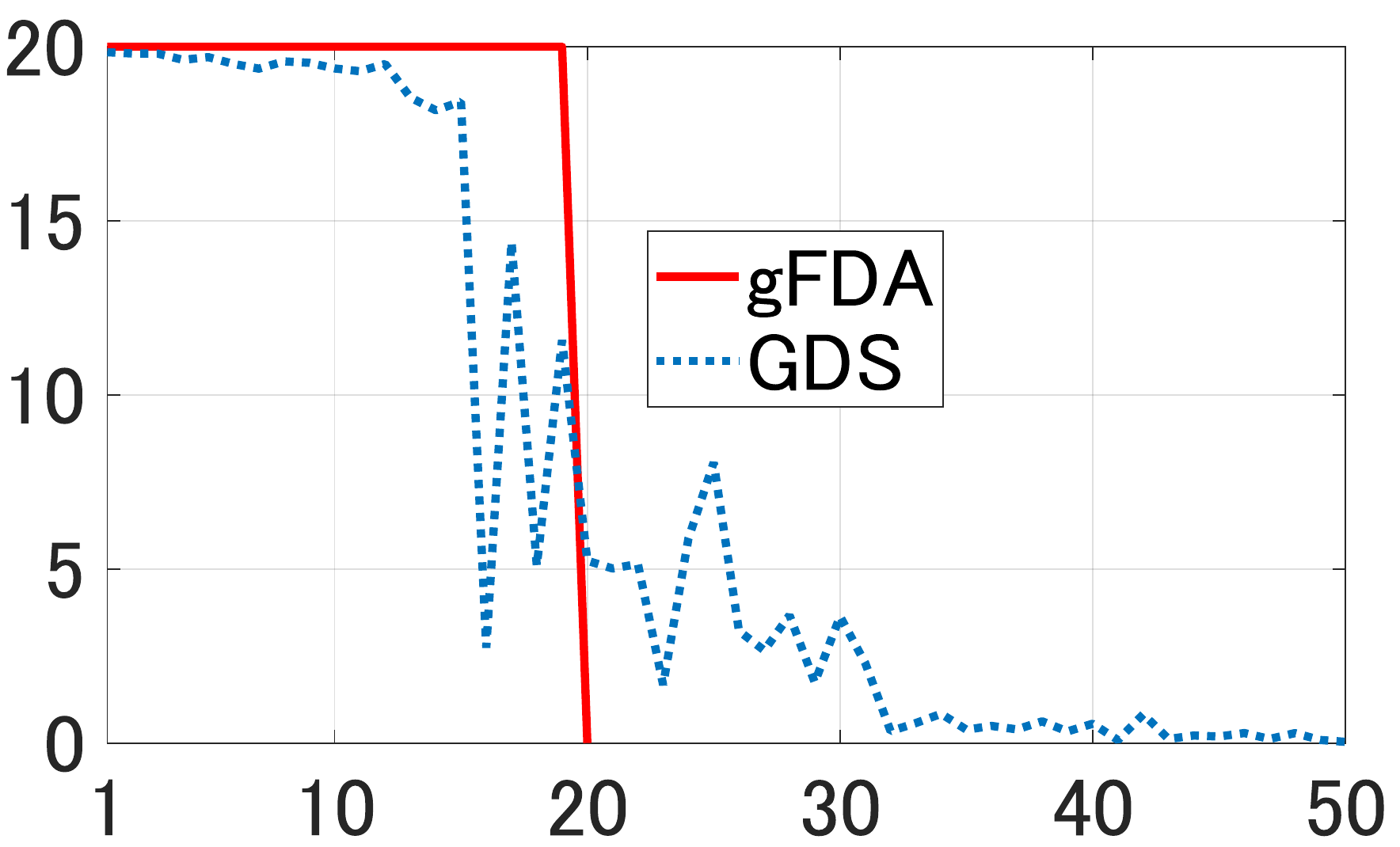}}\\
\subfloat[Eigenvalue (100classes)]{\includegraphics[width=43mm]{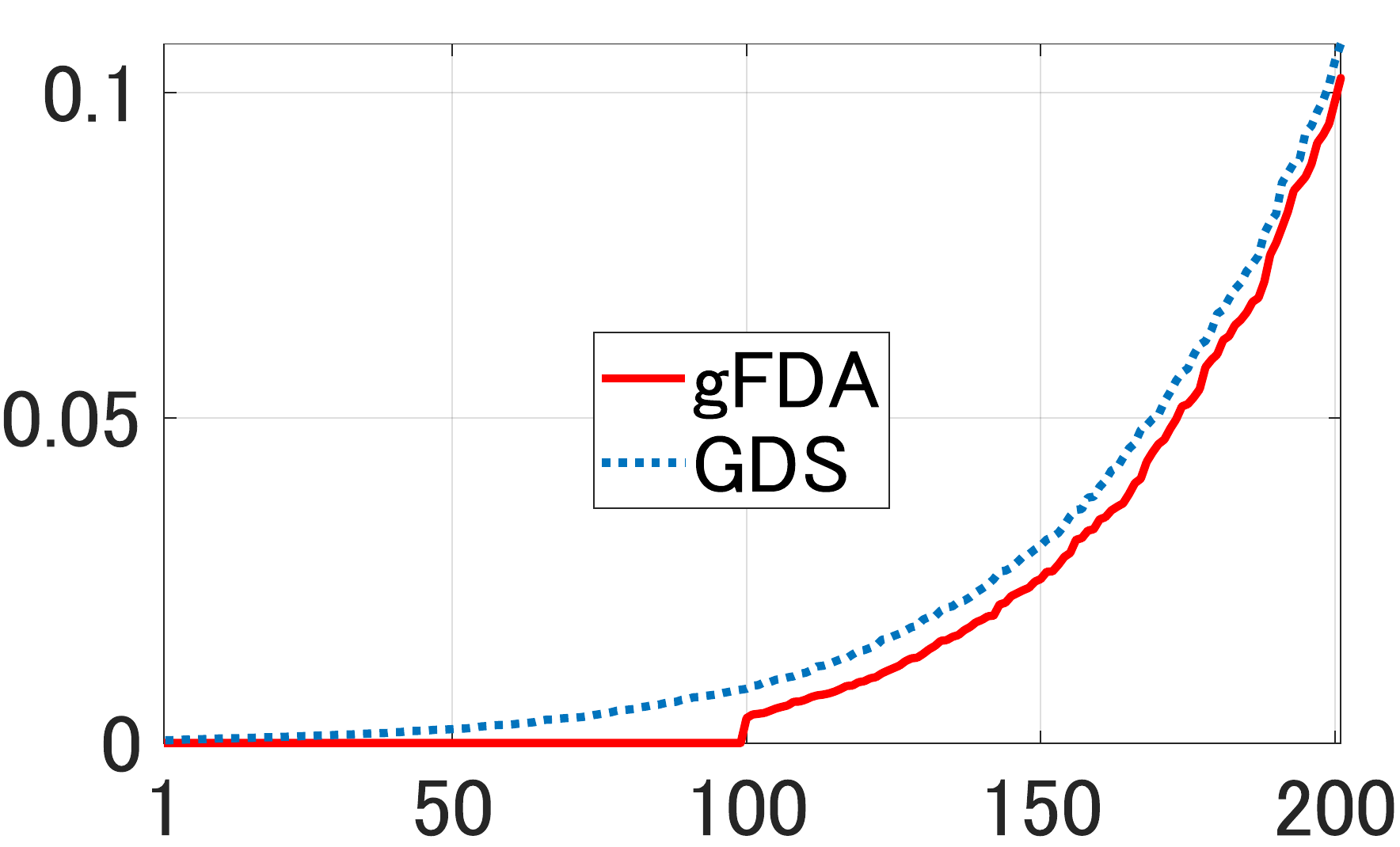}}
\subfloat[F-criterion (100classes)]{\includegraphics[width=43mm]{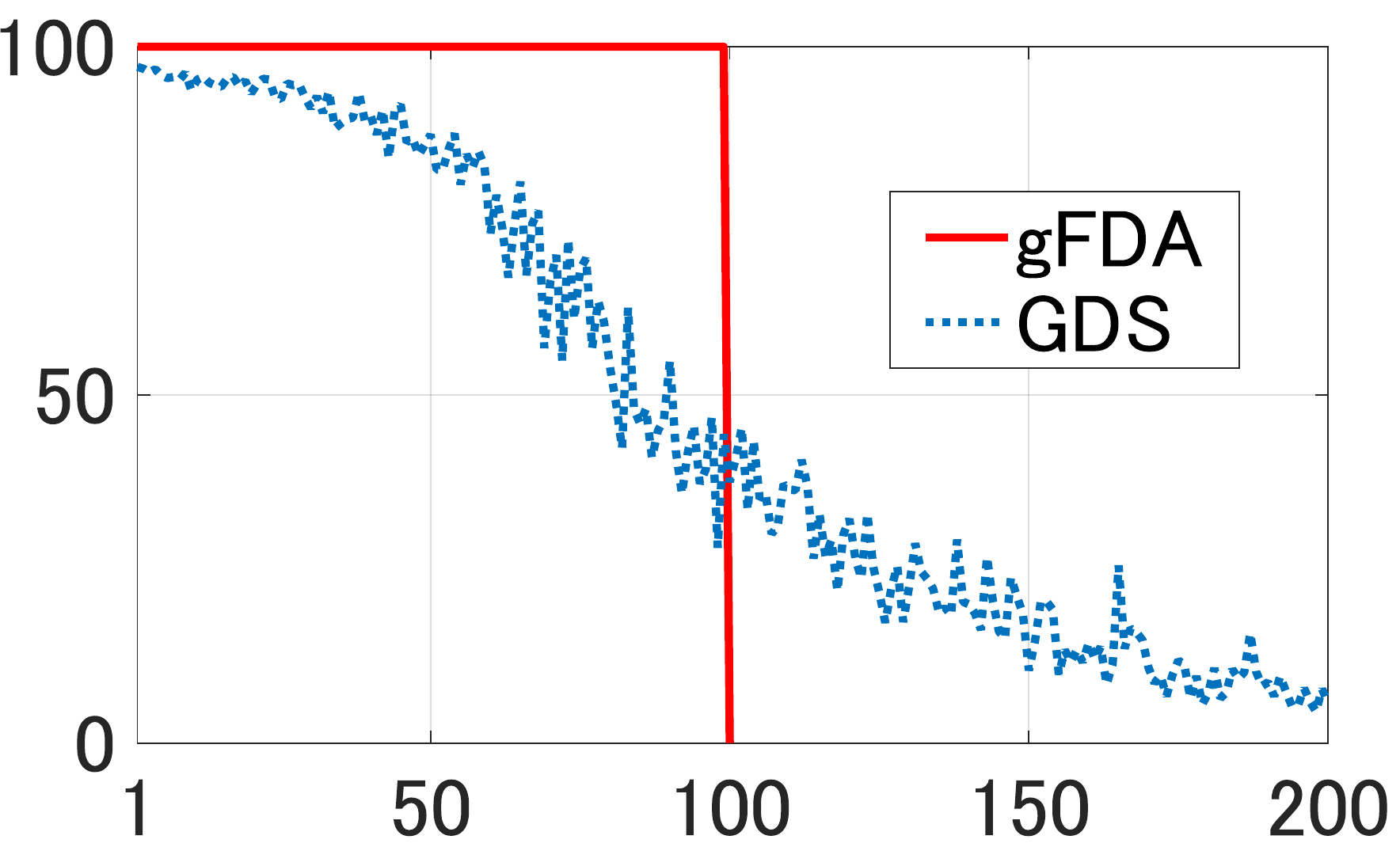}}
\caption{Comparison of gFDA and GDS projection in terms of the distribution of eigenvalues and our Fisher-like criterion.}
\label{comp_gfda_gds_eigens}
  \end{center}
\end{figure}

To show this characteristic more clearly, we compared the distributions of eigenvalues of $\bf{G}$ and $\hat{\bf{G}}$, which were generated from a set of $C$ 3-dimensional class subspaces.
The left column of Fig.\ref{comp_gfda_gds_eigens} shows the eigenvalues from  $\bf{G}$ and $\hat{\bf{G}}$ in the ascending order.
We can see that the $C-1$ smallest eigenvalues are zero for gFDA as described previously. The eigenvalue distributions in both cases are almost the same when the number of classes, $C$, is small as three or five classes. 
However, we can see that the difference between the two distributions becomes larger as the number of class increases. In conjunction with this observation, the value of $\sigma$ also increases from 0.33 (3 classes) to 0.98 (100 classes).

We also compared the gap between these two methods by using another index.
The right column of Fig. \ref{comp_gfda_gds_eigens} indicates the discriminant power of each orthogonal basis vector, which is calculated in terms of our Fisher-like criterion defined in Eq.(\ref{ourCriteria}). In gFDA, the discriminant power of all the basis vectors are equally $C$ as expected from its definition. Thus, the total discriminant power of gFDA is $C \times (C-1)$. For instance, the discriminant power of gFDA is 6 (=3${\times}$2) in the case of 3 classes. In the same sense, the discriminant power of GDS with $C-1$ dimensions can also be measured by the area under the curve over the interval from 1 to $C-1$.
We can see that the discriminant powers of  gFDA and GDS projection are almost equivalent when the class number $C$ is small. 
However, as $C$ becomes larger, the gap between them gets larger, that is, the discriminant power of GDS decreases.

This indicates that the discriminant power of GDS projection could be insufficient when the dimension of GDS is set to $C-1$ as in gFDA, especially in the case with a large number of classes. To deal with this issue, we propose to use a larger dimension than $C-1$ for GDS. Concretely, we take the eigenvectors corresponding to $N_d$ smallest eigenvalues such that the total sum of the discriminant power over the $N_d$ eigenvectors gets larger than a specified threshold value, $\beta = C(C-1){\times}\gamma$, where $\gamma$ is empirically set to a value in the range between 0.8 and 0.95. 

\section{Evaluation experiments}\label{s:experiments}
In this section, we first verify the validity of our heuristic principle on real data.
We then evaluate the effectiveness of gFDA and GDS projection from the following aspects:  1) Fisher-like discriminant ability, 2) their robustness against small variations in training data and large bias between data for training and testing, and 3) classification performance. 

\subsection{Validity of our heuristic principle}
We verify the validity of the heuristic principle: the equivalence in terms of direction between the first orthonormal basis vector ${\bs{\phi}}_1$ and the mean vector ${\bf{m}}$ of a class distribution. For this purpose, we measured the normalized correlation coefficient between ${\bs{\phi}}_1$ and ${\bf{m}}$ in each set of 9PL images of 29 subjects from the Yale face database B+, which were used in Sec.\ref{compFDAandgFDA}. The average of the correlation coefficients of all the subjects was 0.99932. For the CMU face database, which will be used for evaluation later, the average value of 120 subject classes was 0.99998, where each class consists of 20 frontal face images under different illuminations. These high correlations support the validity of our heuristic principle.

Hence, we expect ${\SB}_2$ to be practically almost the same as ${\SB}$ and the simplified matrix ${\SB}_3$ to still properly approximate the original ${\SB}$. In fact, the similarities between ${\SB}_3$ and the original ${\SB}$ were 0.943 and 0.964, in the cases of two and three subject classes, respectively. In this measurement, we first vectorized the covariance matrices and then calculated the normalized correlation coefficient between them. 

In addition, we measured the degree of coincidence between the discriminant spaces that were generated by the original FDA, sFDA with $\frac{{\bf{d}}^T{\SB}_2{\bf{d}}}{{\bf{d}}^T{\SW}_2{\bf{d}}}$, and gFDA with $\frac{{\bf{d}}^T{\SB}_3{\bf{d}}}{{\bf{d}}^T{\SW}_4{\bf{d}}}$.
As the original FDA cannot work on our experimental setting due to the small sample size problem, we used regLDA \cite{regFDA} instead.
Table \ref{compdspace} shows the cosines of the canonical angles between them. We can see that gFDA can be still regarded as a reasonable approximation of the original FDA despite the considerable simplification of the original Fisher criterion.

\begin{table}[bt]
\caption{Comparison of the three discriminant spaces by FDA, sFDA and gFDA.}
\label{compdspace}
\begin{center}
\begin{tabular}{|c|c|c|c|}
\hline
\multicolumn{2}{|c}{2 classes} & \multicolumn{2}{|c|}{3 classes}\\
\hline
sFDA{$\leftrightarrow$}FDA &  gFDA{$\leftrightarrow$}FDA & sFDA{$\leftrightarrow$}FDA & gFDA{$\leftrightarrow$}FDA \\
\hline
$\cos \theta_1$=0.988 & $\cos \theta_1$=0.937 & $\cos \theta_1$=0.970 & $\cos \theta_1$=0.904\\ 
- & - & $ \cos \theta_2$=0.904 & $\cos \theta_2$=0.888 \\
\hline
\end{tabular}
\end{center}
\end{table}

\subsection{Fisher-like discriminant ability}
We verify that gFDA and GDS projection have inherited the high discriminant ability from FDA on the Yale face database B+.

\vspace{1mm}
\noindent{\bf{Experimental settings}}:~
The Yale face database B+ consists of face images of 38 subjects, where these images were acquired under 64 different lighting conditions in nine different poses~\cite{illuminationCone}.
We selected 29 individuals from the database; these individuals' images appear across the four subsets. In the evaluation, we used only the frontal face images, so that our data set contains 1,035 images of 29 subjects under 45 different lighting conditions. We converted the cropped images of 640 \(\times\) 480 pixels to images of 32 \(\times\) 24 pixels and normalized the image vectors.

We conducted evaluation experiments on this database.
The 9PL images of each subject, which were described in Sec.\ref{compFDAandgFDA}, were used for learning and the remaining 36 images were used for testing. 
To verify the robustness of the methods 
against few sample data, 
we changed the number of training data from two to nine, where they were randomly selected from the nine 9PL images.
We repeated this sampling 60 times and calculated the averages of all the results obtained as the final one. 

We conducted experiments under the above condition in the cases of two and 29 classes, using a nearest neighbor classifier with the square of L2 distance between an input data and each class mean. In the case of two classes, we randomly selected 25 pairs of two classes from 29 subject classes and used the average results as the final performance. 
We evaluated the performances of the methods in terms of the recognition rate (\%) and equal error rate (EER) (\%). 

We used three typical variants of modified FDA: pcaLDA \cite{fda2}, nullLDA \cite{newLDA} and regLDA \cite{regFDA}, since the original FDA cannot work on this experimental setting due to the SSS problem. In the sequel, we will refer these modified LDAs as original LDAs for simplicity. We also evaluated the performances of gFDA and GDS with the normalization, which we denote as gFDA+N and GDS+N, respectively. 
The dimension of class subspace of gFDA and GDS projection was set to the number of training data of each class. For GDS projection, the value of $\gamma$ was set to 0.90 to determine the dimension of GDS. For pcaLDA, the sum of squared residuals in PCA used for dimension reduction were 1e-2 and 1e-9 for two and 29 classes, respectively. The value of $\delta$ in regLDA was set to 1e-4 for both. The parameters were empirically determined using the training data set.



\begin{figure}[tb]
\begin{center}
\subfloat[2 classes]{\includegraphics[width=95mm]{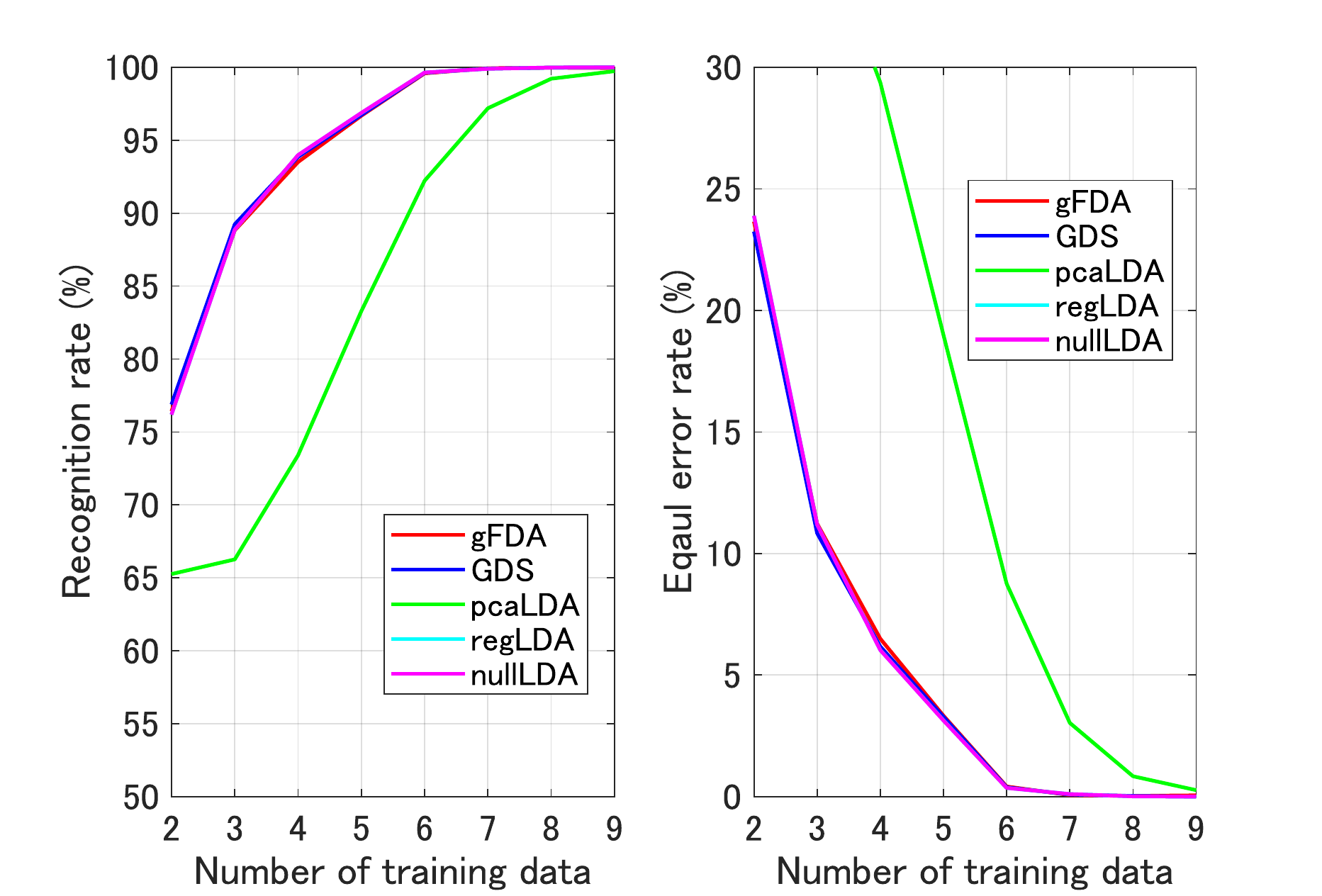}}\\
\subfloat[29 classes]{\includegraphics[width=95mm]{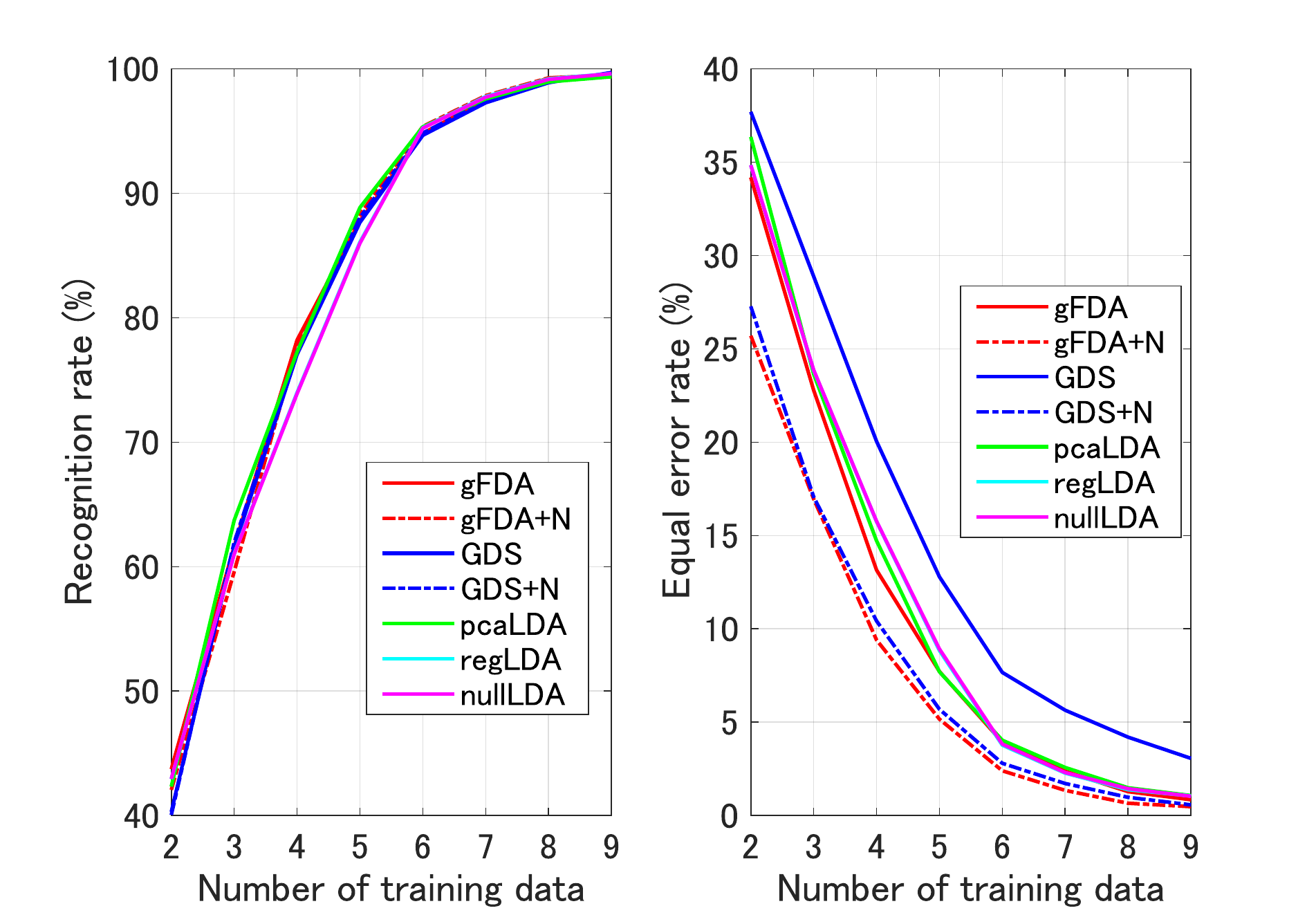}}
\caption{Discriminant ability of the different methods on the Yale face database in terms of recognition rate and equal error rate (\%) in the cases of 2 and 29 classes.}
\label{comp_ability}
\end{center}
\end{figure}

\vspace{1mm}
\noindent{\bf{Experimental results and consideration}}: Figs.\ref{comp_ability}a and b show the results of the different methods in the case of two and 29 classes, respectively. In the figures, the horizontal axis indicates the number of training data and the vertical axes in the left and right panels indicate the recognition rate (\%) and EER (equal error rate) (\%), respectively. 

In the case of two classes, the performances of gFDA and GDS projection are almost the same as those of regLDA and nullLDA, as shown in Fig.\ref{comp_ability}a. This supports that gFDA certainly inherits the discriminant ability of the original FDA and furthermore GDS projection inherits the discriminant ability from gFDA.
Thanks to the characteristic of the illumination subspaces generated from the 9PL images, the recognition rates of all the methods were almost perfect when using all the 9PL images as training data. 

The performance of pcaLDA is very low compared with the other methods, especially when the number of training data is small.
This can be ascribed to the fact that the dimension reduction based on the PCA could not estimate a meaningful within-class covariance from very few training data. For example, only six training data were used for conducting PCA when the number of training data is three for each class.

The close relationship among the methods retains in the case of 29 classes as shown in Fig.\ref{comp_ability}b. Overall, gFDA+N and GDS+N significantly outperform the other methods as we expected. Although the performance of GDS projection was lower than those of the other methods in terms of EER, it has been visibly improved by the normalization. 

\begin{figure}[tb]
\begin{center}
\subfloat[Set1: with small variance]{\includegraphics[width=40mm]{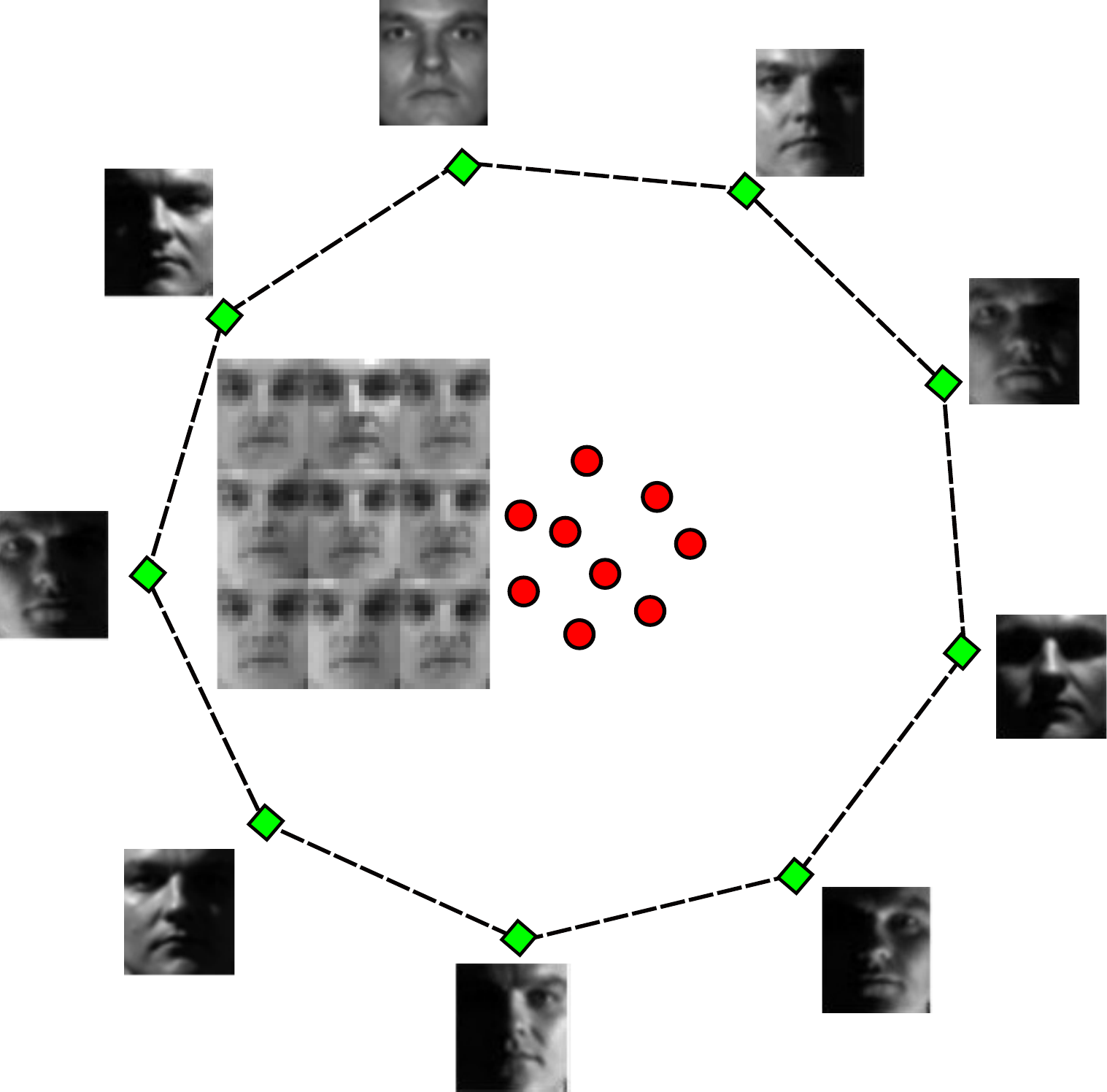}}\hspace{3mm}
\subfloat[Set2: with large bias]{\includegraphics[width=40mm]{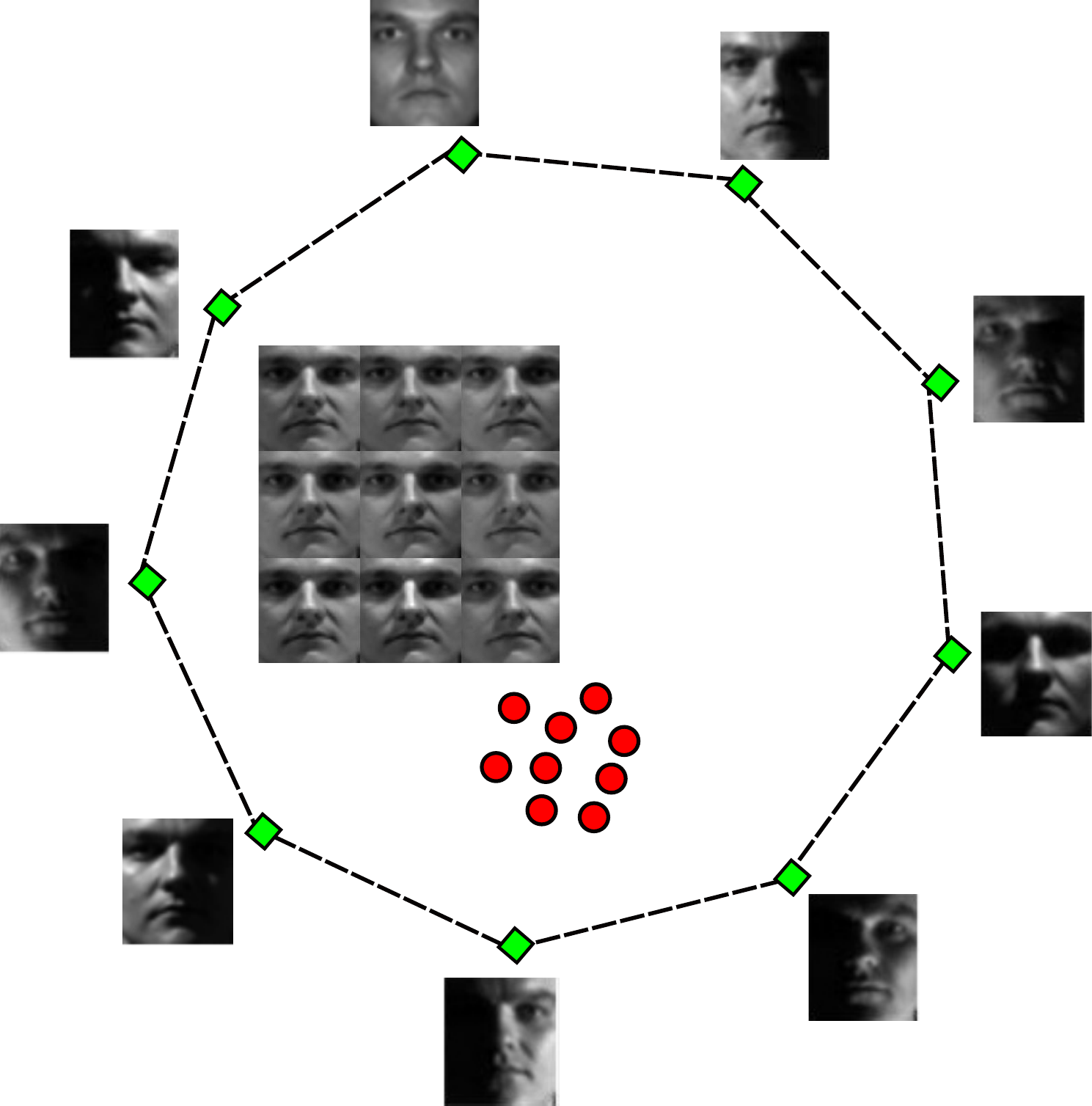}}
\caption{Conceptual illustration of training sets: the 9PL images and two types of the sets of face images generated from the 9PL images. The 9PL images and the generated images are indicated by green diamonds and red circles, respectively.}
\label{learning_images}
\end{center}
\end{figure}

\subsection{Robustness against small variation and large bias}
We evaluated the robustness of gFDA and GDS projection against small data variation in the training data of a class and large bias between the training and test data in a class.

\vspace{1mm}
\noindent{\bf{Experimental settings}}: For the evaluation, we synthesized two types of data sets, Set1 and Set2, from the 9PL images.  

\noindent {\bf{Set1}}: a set of nine face images $\{\bf{S}\}$, which were randomly synthesized as linear combinations of the 9PL images, ${\{{\bf{L}}_i\}}_{i=1}^9$, with the nonnegative constraint, such that the synthesized nine images $\{\bf{S}\}$ are distributed with small variation around the mean face image ${\bf{L}}_m$ as shown in Fig. \ref{learning_images}a. More concretely, we generated a set of $\{\bf{S}\}$ by repeating the following two steps: 1) random sampling: ${\bf{S'}}=5 {\times}{\bf{L}}_m  + \sum_{i=1}^9 {c_i}{\bf{L}}_i$, where $c_i$ is positive random variable with uniform distribution subject to $\sum_{i=1}^9 c_i =1$; and 2) normalization:  ${\bf{S}}={\bf{S}'}/{||\bf{S}'||}$. We generated 60 data sets by conducting these synthesis processes 60 times.

\noindent {\bf{Set2}}: a set of nine face images, which were randomly synthesized from the 9PL images such that the synthesized images are distributed with biases from the mean image as shown in Fig.\ref{learning_images}b. We generated a set of $\{\bf{S}\}$ by repeating the following two steps: 1) random sampling: ${\bf{S'}}=2{\times}{\bf{L}}_j  + \sum_{\substack{i=1 \\ i \neq j}}^9 {c_i}{\bf{L}}_i$, where $j$ is a number selected randomly in the range from one to nine and $c_i$ is positive random variable with uniform distribution subject to $\sum_{i=1}^9 c_i =1$; and  2) normalization:  ${\bf{S}}={\bf{S}'}/{||\bf{S}'||}.$  By repeating this generation, we generated 60 sets of this distribution type.
We used a nearest neighbor classifier based on class means. 
For pcaLDA, the cumulative energy of PCA were 1e-9 and 1e-6 for set1 and set, respectively. For the other parameters for all the methods, we used 
the same values as that used in the previous experiment. 

\begin{figure}[tb]
\begin{center}
\subfloat[Set1]{\includegraphics[width=90mm]{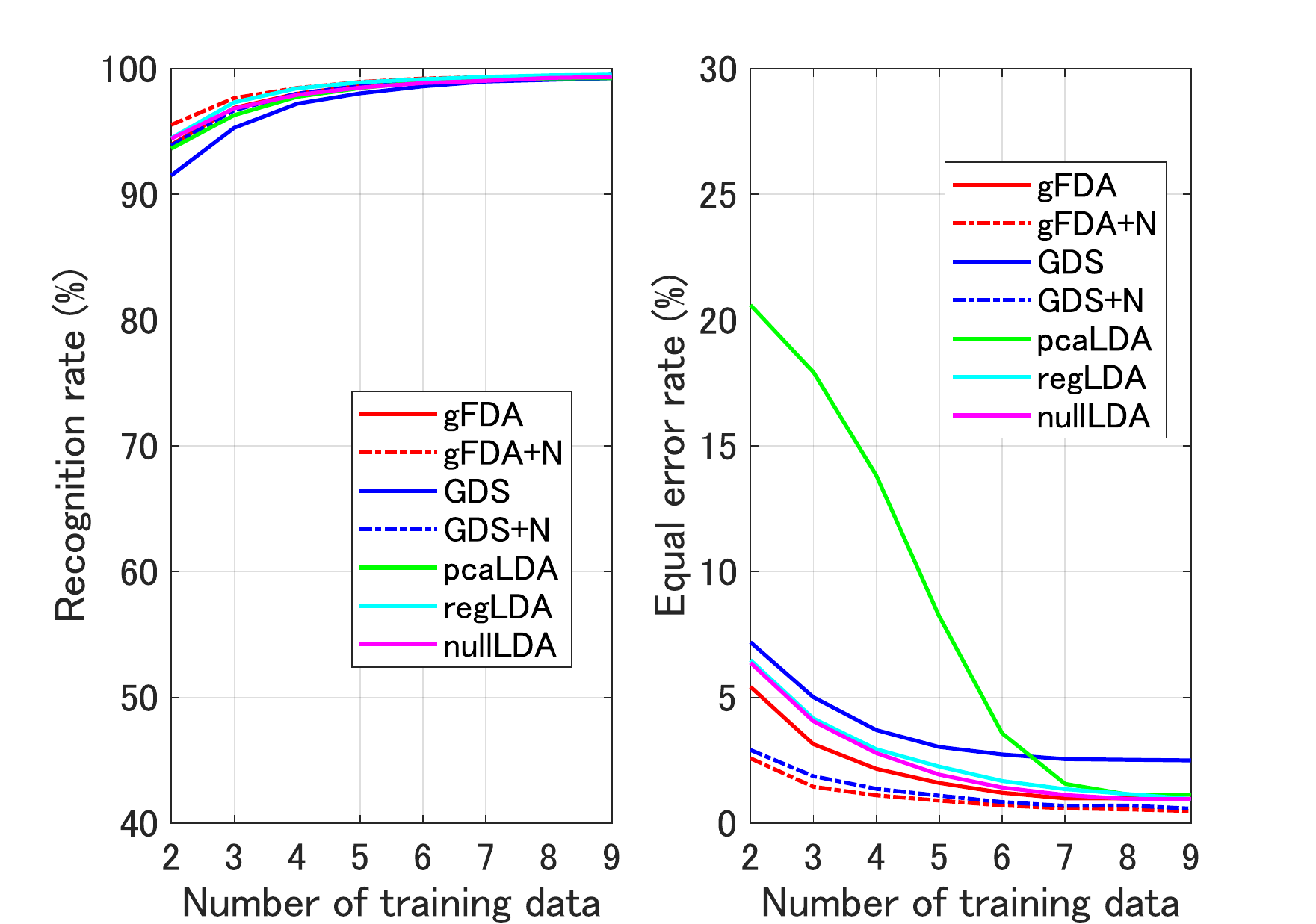}}\\
\subfloat[Set2]{\includegraphics[width=90mm]{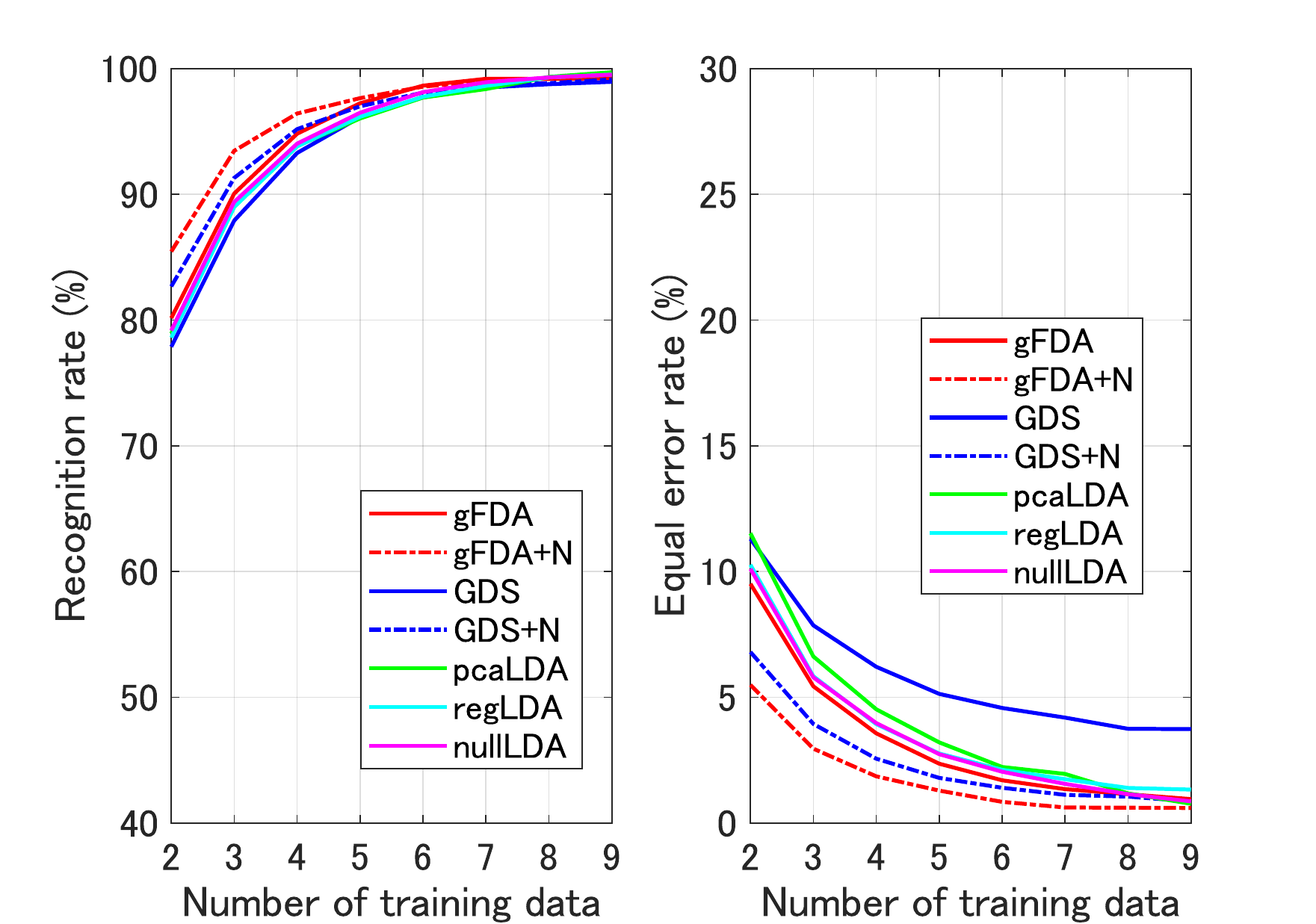}}\\

\caption{Comparison of different methods in terms of recognition rate and equal error rate (\%) in 29 classes.}
\label{comprobustness}
\end{center}
\end{figure}

\vspace{1mm}
\noindent{\bf{Experimental results and consideration}}: Figs.\ref{comprobustness}a and b show the results of the different methods on Set1 and Set2. 
Overall, gFDA and the original LDAs have the same level of performance.
However, when we look at the details, we notice the advantage of gFDA against the original LDAs. In particular, the EER of gFDA is very low when the number of training data is small (from three to five).
This is because the illumination subspace corresponding to ${\SW}_4$ could be stably generated even from few training data, as the data of Set1 and Set2 are contained in the illumination subspace. Moreover, the normalization significantly improved the performance of gFDA.
The performance of GDS is not so high in comparison with that of the gFDA. This is because the geometrical difference between gFDA and GDS might be much larger than anticipated, under the situation with a large data variation. However, the performance of GDS has been largely improved by the normalization 
and achieved the same level as gFDA. 
Overall, the performances of the gFDA+N and GDS+N are especially prominent compared with the other methods.

\begin{figure}[tb]
\begin{center}
\includegraphics[width=95mm]{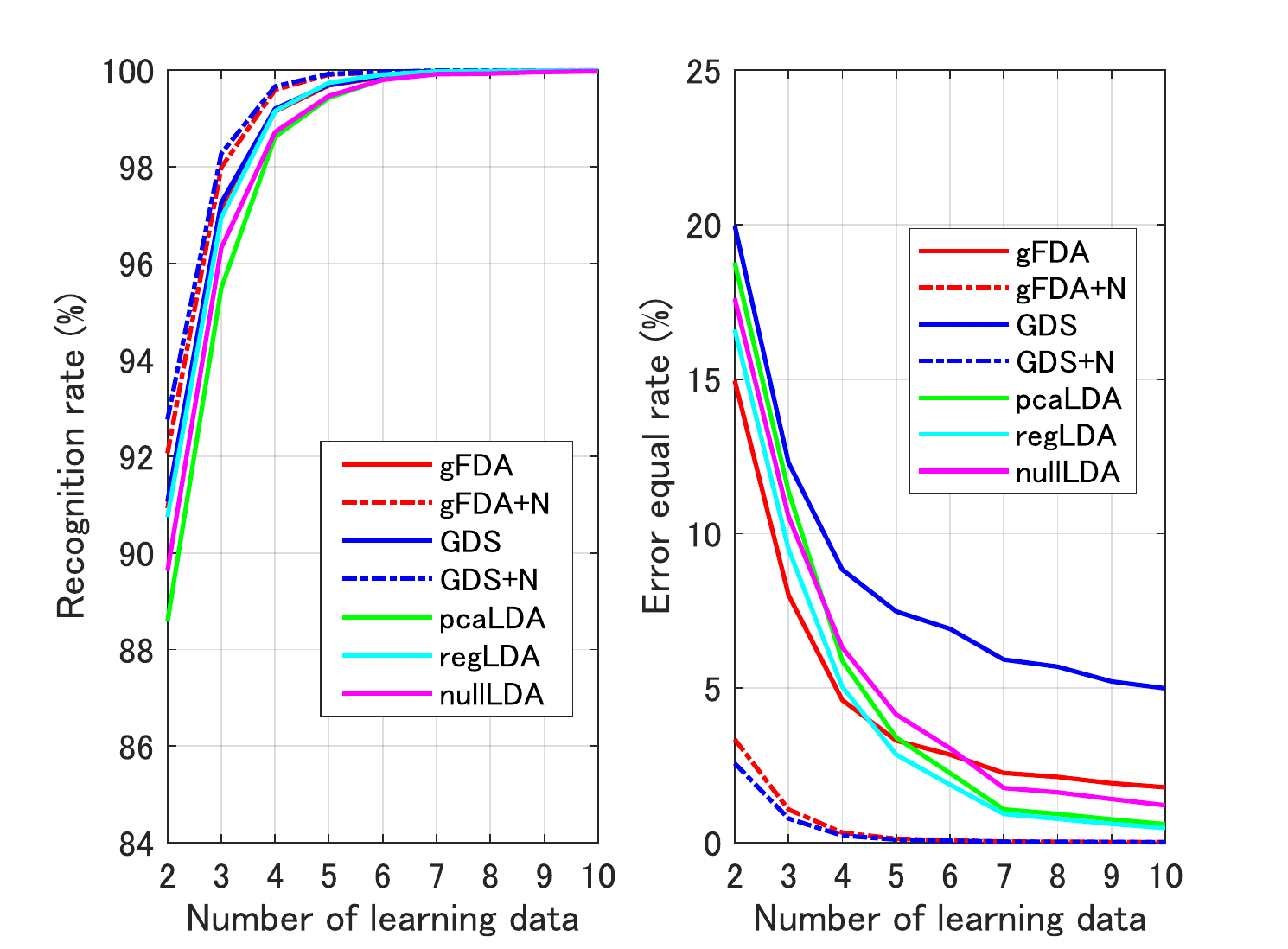}
\caption{Comparison of different methods in terms of recognition rate and equal error rate (\%) in 128 classes from the CMU face database.}
\label{comp2_128classes}
\end{center}
\end{figure}

\subsection{Evaluation of classification performance}
We conducted the classification on larger scale data from the CMU Multi-PIE face database.

\vspace{1mm}
\noindent{\bf{Experimental settings}}:~
The CMU Multi-PIE face database consists of face images of 337 subjects, captured from 15 viewpoints with 20 different lighting conditions in four recording sessions \cite{mpie}. In the experiment, we used frontal face images of 128 subjects across all four sessions. We took a sub-sampled image of size 36$\times$36 pixels from an original image, where we cropped this image by reference to the two inner corners of the eyes and the tip of the nose. The vectorized images were normalized. In classification, we used a nearest neighbor classifier with the square of L2 distance between the input and each class mean.

For each subject on a session, $n$ images 
sampled randomly from 20 images were used for training and the remaining 20-$n$ images 
for testing. We evaluated the performance of the methods while increasing the number of training data, $n$, from two to ten. We repeated this evaluation 60 times for each $n$ and then calculated the average of the results. Further, we conducted the same evaluation on the three remaining sessions and took the average of the results on the four sessions as a final recognition performance. For gFDA and GDS projection, we set the dimension of class subspace to the number of training data, $n$. For GDS, the value of $\gamma$ was set to 0.90.
The sum of squared residuals of PCA used in pcaLDA was 2e-4. For regLDA, $\delta$ was 1e-4. The other parameters were set to the same 
values as those used 
in the previous experiment. 

\vspace{1mm}
\noindent{\bf{Experimental results and consideration}}: Fig.\ref{comp2_128classes} shows the classification performances of the different methods in terms of recognition rate and equal error rate (EER). 
The overall trend remains 
the almost same as
the previous experiments: gFDA is comparatively superior to the FDAs when the number of training data is small (from two to four), gFDA has the same level in its performance as those of 
the FDAs when $n$ is large (over 5), 
and GDS projection is poorer than the other methods particularly in terms of EER. However, the effectiveness of the normalization in this case is much clearer in comparison with those in the previous cases; gFDA+N and GDS+N significantly outperform the other methods in both indexes. This result supports that the normalization is intrinsically required to get the best performance out of gFDA and GDS projection. Moreover, in the extreme case that only one learning data is available, pcaLDA and nullLDA cannot work in principle, but gFDA+N and GDS+N can still work with the recognition rates of 53.2\% and 47.8\%, and the EERs of 15.9\% and 17.1\%, respectively.

We discuss the validity of subspace representation in gFDA and GDS under the SSS problem. The variation of illumination conditions on the CMU database is smaller than that on the Yale database, which leads that 20 images of a subject can be contained within a 5-dimensional illumination subspace with the sum of squared residual error of 0.1\%. This setting enables gFDA and GDS to reach almost perfect performance with only five training images, while nine training images were needed in the previous experiment on the Yale database.

\section{Conclusion}\label{s:conclusion}
In this paper, we proposed a new type of discriminant analysis based on the orthogonal projection of data onto a generalized difference subspace (GDS).
We revealed that the GDS projection functions as a discriminant feature extraction through a similar mechanism as the Fisher discriminant analysis (FDA).
In this process, we introduced geometrical Fisher discriminant analysis (gFDA), which is a discriminant analysis based on a simplified Fisher criterion.
We then proved that gFDA is equivalent to GDS projection with a small correction term. This equivalence ensures GDS projection to inherit the discriminant ability from FDA by regarding gFDA as an intermediate concept between them. 
To further enhance the performances of gFDA and GDS projection, we proposed to normalize the projected vectors onto the discriminant spaces. 
Extensive experiments using the extended Yale B+ database and the CMU face database showed that gFDA and GDS projection have high discriminant ability as well as FDA, and their extensions with normalization have equivalent or higher performance than conventional FDAs.


\ifCLASSOPTIONcompsoc
  \section*{Acknowledgments}
This work was partly supported by JSPS KAKENHI Grant Number 16H02842.
We would like to thank Hideitsu Hino and Rui Zhu for their helpful comments.
\else
  \section*{Acknowledgment}
\fi

\ifCLASSOPTIONcaptionsoff
  \newpage
\fi

\bibliographystyle{IEEEtran}
\bibliography{references.bib}

\end{document}